\pdfoutput=1
\pdftrailerid{}
\documentclass[11pt]{article}

\usepackage[preprint]{acl}
\usepackage{times}
\usepackage{latexsym}
\usepackage[T1]{fontenc}
\usepackage[utf8]{inputenc}
\usepackage{microtype}
\usepackage{graphicx}
\usepackage{amsmath,amssymb,amsthm,mathtools}
\usepackage{booktabs,multirow}
\usepackage{algorithm}
\usepackage[noend]{algpseudocode}
\usepackage{xcolor}
\usepackage{enumitem}
\usepackage[capitalise,noabbrev]{cleveref}

\newcommand{\Htok}{H_{\mathrm{tok}}}
\newcommand{\Hstr}{H_{\mathrm{str}}}
\newcommand{\Hhom}{H_{\mathrm{hom}}}
\newcommand{\Hroot}{H_{\mathrm{root}}}
\newcommand{\Hbranch}{H_{\mathrm{branch}}}
\newcommand{\Hstop}{H_{\mathrm{stop}}}
\newcommand{\Pt}{\tilde{P}}          
\newcommand{\supp}{\operatorname{supp}}
\newcommand{\ch}{\operatorname{ch}}  
\newcommand{\Roots}{\mathcal{R}}
\newcommand{\Voc}{\mathcal{V}}
\newcommand{\bytes}{\beta}
\newcommand{\hb}{h_2}                
\newcommand{\E}{\mathbb{E}}
\newcommand{\1}{\mathbf{1}}
\newcommand{\KL}{\mathrm{KL}}
\newcommand{\pfx}{\prec}             
\newcommand{\pfxeq}{\preceq}
\newcommand{\spbar}{\rule{0.42em}{0.14ex}} 

\definecolor{cRoot}{HTML}{0072B2}
\definecolor{cBranch}{HTML}{009E73}
\definecolor{cStop}{HTML}{D55E00}
\definecolor{cHom}{HTML}{7F7F7F}
\definecolor{cEN}{HTML}{000000}   
\definecolor{cZH}{HTML}{E69F00}
\definecolor{cJA}{HTML}{CC79A7}

\InputIfFileExists{gen/numbers}{}{\newcommand{\res}[4]{\textcolor{red}{??}}}
\InputIfFileExists{gen/numbers_extra}{}{}   
\InputIfFileExists{gen/numbers_rts}{}{}     

\newtheorem{theorem}{Theorem}
\newtheorem{proposition}{Proposition}
\newtheorem{lemma}{Lemma}
\newtheorem{corollary}{Corollary}
\theoremstyle{definition}
\newtheorem{definition}{Definition}
\theoremstyle{remark}
\newtheorem{remark}{Remark}

\title{Optimal Single-History Tokenization Disambiguation\\
for Provably Secure Linguistic Steganography}

\author{Feng Jiang\thanks{\url{https://github.com/lxjf12138/coin-projection}} \\
  \texttt{fengjiang@hdu.edu.cn}}

\begin{document}
\maketitle

\begin{abstract}
Provably secure linguistic steganography assumes the receiver recovers the sender's tokens, but the channel carries only text, which can have several tokenizations.
Existing fixes bias sampling, lose most capacity at large vocabularies, run the model on several tokenizations per step, or repair bit errors over a second text.
We prove that for any universal scheme whose sender queries the language model only along the emitted token history, exactness and zero error across models force each key's candidate strings to be prefix-free at every step, however far ahead the receiver reads.
Each step therefore loses at least the stop entropy, the graph entropy of deciding whether a token ends at a string that other candidates extend.
Our scheme, coin projection, makes these stop decisions with shared randomness at one model call per token; it is exact and attains the bound.
On English, Chinese and Japanese, at most \res{all}{amb}{any}{max} in 1,000 sampled tokens are tokenized non-canonically, yet the stop entropy takes \res{all}{e0}{any}{stopmin}--\res{all}{e0}{any}{stopmax}\% of the string entropy.
Coin projection turns \res{all}{gain}{Full}{min}--\res{all}{gain}{Full}{max} times as much entropy into message bits as SyncPool at the full vocabulary, and carries \res{all}{bestk}{syncpool}{min}--\res{all}{bestk}{syncpool}{max} times its bits per token at each method's best vocabulary size.
It decoded all \res{all}{recvall}{coin}{decoded} texts of the main comparison and, at the full vocabulary, came within \res{all}{las}{Full}{utilgapmin}--\res{all}{las}{Full}{utilgapmax} utilization points of a multi-history scheme making \res{qwen}{time@las}{Full}{fwdpertok}--\res{llama}{time@las}{Full}{fwdpertok} calls per token.
\end{abstract}

\section{Introduction}
\label{sec:intro}

Provably secure linguistic steganography hides a message in text sampled from a language model.
The stegotext then follows the model's own distribution, and no warden without the key can tell it from ordinary output \citep{hopper2002provably,ziegler2019neural,kaptchuk2021meteor,de2022perfectly,ding2023discop}.
Without building in a specific model, these universal schemes \citep{dedic2009upper} query the model only along the emitted token history and assume that the receiver recovers the sender's tokens.
Real channels carry text, however, and one text admits many tokenizations.
When the receiver retokenizes, it can recover a different sequence; with 7B models, 5--10\% of 128-bit messages are lost \citep{yan2025addressing}.

Each existing fix pays for disambiguation.
\emph{Filters} remove candidates so that the text parses uniquely \citep{nozaki2022addressing,yan2023secure,yan2025addressing}.
They thereby change the distribution that security rests on.
\emph{SyncPool} keeps the distribution exact by encoding pools of prefix-related tokens \citep{qi2024provably}.
A single short token, however, can be a byte prefix of most candidates; the bare space is a prefix of nearly every English word token.
The pools then merge, and capacity collapses as the vocabulary grows \citep{yan2025addressing}.
On Llama-3.1-8B with the full vocabulary, SyncPool's coder sees only \res{llama}{e0}{Full}{root}\% of the string entropy (\cref{sec:cost}).
\emph{Multi-history} schemes such as LAS \citep{feng2025high} code nearly all the entropy but call the model on several tokenizations.
ReTokSync \citep{wang2026retoksync} repairs bit errors over a second text.

\begin{figure}[t]
\centering
\includegraphics[width=\columnwidth]{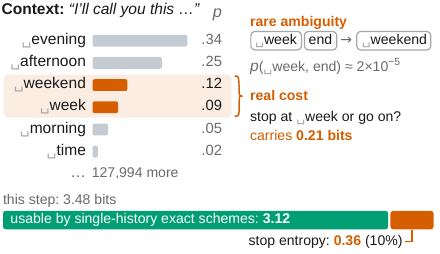}
\caption{The ambiguous outcome is rare, yet a single-history scheme pays for every stop decision.
Top next tokens of Llama-3.1-8B (full vocabulary) after ``I'll call you this''; \textvisiblespace{} marks a space.
If the sender draws \texttt{\textvisiblespace week} and then \texttt{end} (probability $2\times10^{-5}$), the receiver reads \texttt{\textvisiblespace weekend} and loses the message.
Deciding whether the token ends at \texttt{\textvisiblespace week} carries 0.21 bits; all such decisions carry $\Hstop=0.36$ of the step's 3.48 bits, which no single-history scheme that is exact and zero-error for every model can embed (\cref{thm:bound}).}
\label{fig:example}
\end{figure}

We ask how much capacity a scheme must give up if it is exact and zero-error for every model and calls the model only along the emitted token history, as SyncPool and ordinary sampling do.
The answer is the \emph{stop entropy} $\Hstop$.
Whenever a candidate string is a byte prefix of other candidates, the scheme must decide at this step whether the token ends there, and if the message makes this decision, no receiver can recover it from the text, however far ahead the receiver reads (\cref{fig:example}).
Our scheme, \emph{coin projection}, lets the shared key make every such stop decision and leaves the rest of the choice, which prefix tree and which branch, to the message.
For each key, the strings that remain are prefix-free, so the receiver parses the text without lookahead; averaged over keys, every string keeps its probability, so the text still follows ordinary sampling.

Our contributions are as follows.
\begin{itemize}[nosep,leftmargin=*]
\item \textbf{A tight bound.}
In every single-history scheme that is exact and zero-error for every model, the strings a key allows at each step form a prefix-free set, even though the receiver may read arbitrarily far ahead.
Each step therefore loses at least $\Hstop$, the graph entropy of the prefix comparability graph \citep{korner1973coding}, and in any scheme that attains this bound, only the stop decisions depend on the key (\cref{thm:antichain,thm:bound}).
\item \textbf{An optimal scheme.}
Coin projection attains the bound in closed form; for every key, its coder sees at least SyncPool's entropy (\cref{sec:coin}).
\item \textbf{Where the entropy goes.}
On English, Chinese and Japanese, $\Hstop$ stays at \res{all}{e0}{any}{stopmin}--\res{all}{e0}{any}{stopmax}\% of the string entropy at every vocabulary size. It comes from words, stems and particles that can either end a token or continue, whereas the bare space and single-byte UTF-8 prefixes start large subtrees, rarely end a token, and inflate only the branch entropy that SyncPool gives up (\cref{sec:cost}).
End to end, coin projection's gain over SyncPool matches the prediction from these entropies, and at the full vocabulary it comes within \res{all}{las}{Full}{utilgapmin}--\res{all}{las}{Full}{utilgapmax} utilization points of LAS, which makes \res{qwen}{time@las}{Full}{fwdpertok}--\res{llama}{time@las}{Full}{fwdpertok} calls per token (\cref{sec:experiments}).
\end{itemize}

\section{Background}
\label{sec:prelim}

\paragraph{Provably secure steganography.}
Sender and receiver share a secret key $K$ and a context.
At step $t$, the language model gives a token law $\pi_t$ on the history, truncated to the top $k$ tokens for a public $k$.
A distribution coder takes a law $P$ over a finite alphabet, the state of the message, and pseudo-random bits derived from $K$, and outputs a symbol.
Over the key, the symbol follows $P$ exactly.
Given $P$ and the symbol, the receiver updates its knowledge of the message.
Arithmetic coding \citep{ziegler2019neural}, Meteor \citep{kaptchuk2021meteor}, iMEC \citep{de2022perfectly}, Discop \citep{ding2023discop}, Shimmer \citep{bai2025shimmer}, SparSamp \citep{wang2025sparsamp} and range coding \citep{yan2026efficient} all fit this interface and embed up to $H(P)$ bits per step.
If $P=\pi_t$ at every step, the stegotext follows ordinary sampling, so no warden without the key can tell it from cover text \citep{hopper2002provably,cachin1998information}.

\paragraph{The text channel.}
A text channel, however, delivers only the text.
Each token $x$ spells a non-empty byte string $\bytes(x)$, and different token sequences can spell the same text.
When the receiver retokenizes the text, it obtains the tokenizer's \emph{canonical} tokenization, which may differ from the sequence the sender emitted.
The receiver's coder inputs then differ from the sender's, and the message is lost.

There are two sources of ambiguity.
The first is prefixes, as when candidates \texttt{\textvisiblespace a} and \texttt{\textvisiblespace an} let text starting with \texttt{\textvisiblespace an} parse either way.
The second is \emph{homographs}, distinct tokens that spell the same string, which no text-only receiver can tell apart.
We write $p(v)=\sum_{x:\,\bytes(x)=v}\pi_t(x)$ for the \emph{string law}.
Its entropy is $\Hstr=H(p)$, and $\Htok=H(\pi_t)=\Hstr+\Hhom$, where $\Hhom$ is the expected entropy of the token given its string.

\paragraph{Existing designs.}
Besides \emph{Native}, which codes $\pi_t$ directly, prior schemes fall into three families (\cref{tab:props}).
\emph{Filters} restrict $\pi_t$ to a subset of candidates so that the text parses uniquely.
Basic drops every candidate whose string prefixes another candidate's \citep{nozaki2022addressing}.
MWIS keeps a maximum-weight prefix-free subset \citep{yan2023secure}.
Stepwise verification keeps the candidates whose addition leaves the history canonically tokenized \citep{yan2025addressing}.
Dropping any positive-mass candidate changes $\pi_t$, so the stegotext deviates from ordinary sampling.

\emph{SyncPool} groups prefix-related candidates into pools (the connected components of the prefix relation), encodes the pool, and draws the member with shared randomness \citep{qi2024provably}.
Its law is exact, but the coder sees only pool masses.

\emph{Multi-history} designs evaluate the model on several tokenizations of the text \citep{feng2025high}.

\begin{table}[t]
\centering
\small
\setlength{\tabcolsep}{3.2pt}
\begin{tabular}{@{}lccccc@{}}
\toprule
 & Exact & Zero & Single & Calls & Budget \\
 & law & error & history & /token & /step \\
\midrule
Native & \checkmark & & \checkmark & 1 & $\Htok$ \\
Basic, MWIS & & \checkmark & \checkmark & 1 & --- \\
Stepwise & & \checkmark & \checkmark & 1$^\dagger$ & --- \\
SyncPool & \checkmark & \checkmark & \checkmark & 1 & $\Hroot$ \\
Multi-history & \checkmark & \checkmark & & $>1$ & $\approx\Hstr$ \\
\textbf{Coin (ours)} & \checkmark & \checkmark & \checkmark & 1 & $\Hstr-\Hstop$ \\
\bottomrule
\end{tabular}
\caption{Coin projection is the only single-history design that is exact and zero-error and gives the coder more than the root entropy.
The budget is the expected entropy of the coder input per step (\cref{sec:forest}); filters code a law other than the model's, so they get no budget.
$^\dagger$Plus one retokenization of the history per candidate.}
\label{tab:props}
\end{table}

\section{Where the String Entropy Goes}
\label{sec:cost}
\label{sec:forest}

We fix one generation step and drop the step index.
Let $p$ be the string law at this step (\cref{sec:prelim}) and $S=\supp p$.
Strings are compared byte by byte, and $v\pfx w$ means that $v$ is a proper prefix of $w$.

\begin{definition}[Prefix forest]
\label{def:forest}
The \emph{parent} of $w\in S$ is its longest proper prefix in $S$, if one exists.
This relation makes $S$ a forest.
Its \emph{roots} $\Roots$ are the strings with no proper prefix in $S$, and $\ch(v)$ denotes the children of $v$.
A \emph{node} is a string with at least one child.
The \emph{mass} of $v$ is $M(v)=\sum_{w\in S,\,v\pfxeq w}p(w)$, and the \emph{stop probability} of a node is $q_v=p(v)/M(v)$.
\end{definition}

Drawing $V\sim p$ is the same as walking down the forest.
The walk enters the tree of root $r$ with probability $M(r)$.
At each node $v$ it reaches, it stops with probability $q_v$; otherwise it moves to child $c$ with probability $M(c)/(M(v)-p(v))$.
The chain rule of entropy, applied to this walk, splits the string entropy into three parts.

\begin{lemma}[Grouping]
\label{lem:grouping}
$H(p)=\Hroot+\Hbranch+\Hstop$, where
\begin{align*}
\Hroot &= H\bigl((M(r))_{r\in\Roots}\bigr),\\
\Hstop &= \textstyle\sum_{v\ \mathrm{node}} M(v)\,\hb(q_v),\\
\Hbranch &= \textstyle\sum_{v\ \mathrm{node}} \bigl(M(v)-p(v)\bigr)\,H(\gamma_v),
\end{align*}
$\gamma_v=\bigl(M(c)/(M(v)-p(v))\bigr)_{c\in\ch(v)}$ is the child law at $v$, and $\hb$ is the binary entropy.
\end{lemma}

\begin{figure*}[t]
\centering
\includegraphics[width=\textwidth]{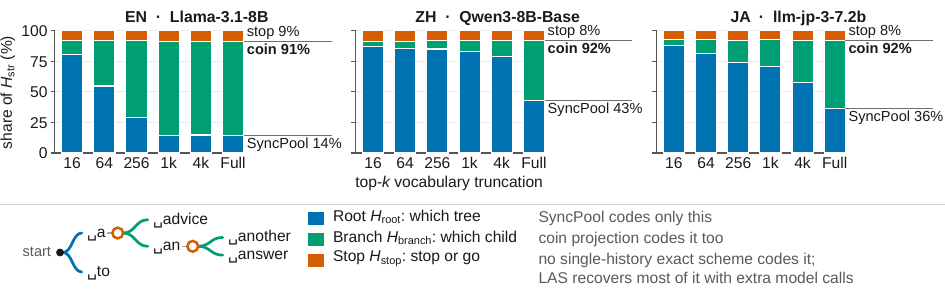}
\caption{Stop entropy stays at \res{all}{e0}{any}{stopmin}--\res{all}{e0}{any}{stopmax}\% of the string entropy at every vocabulary size, while hub tokens collapse the root entropy, the only part SyncPool encodes; coin projection encodes root and branch.
Each bar splits $\Hstr$ of the top-$k$ law by \cref{lem:grouping}, averaged over 3,200 histories of ordinary sampling; $\Hhom<10^{-4}$ bits throughout.
Key: on a small forest, the walk picks a tree, then stops or goes at each node it reaches, and picks a child when it goes.}
\label{fig:decomp}
\end{figure*}

$\Hroot$ is the entropy of choosing a tree, $\Hstop$ the entropy of the stop/go decisions, and $\Hbranch$ the entropy of choosing a child (key of \cref{fig:decomp}).
Together with $\Htok=\Hstr+\Hhom$ (\cref{sec:prelim}), the lemma gives the ladder $\Hroot\le\Hstr-\Hstop\le\Hstr\le\Htok$, and \cref{fig:example} shows $\Hstop$ on one step.

We now measure how much entropy each term carries when Llama-3.1-8B, Qwen3-8B-Base and llm-jp-3-7.2b continue English, Chinese and Japanese C4 texts by ordinary sampling (50 prompts $\times$ 64 tokens per setting; \cref{sec:setup}).

\paragraph{Ambiguous text is rare, yet every step pays the stop entropy.}
In the texts of \cref{sec:experiments}, which follow ordinary sampling, at most \res{all}{amb}{any}{max} in 1,000 tokens differ from canonical retokenization \citep{yan2025addressing} (\cref{app:native-err,app:origin}).
A single-history scheme that is exact and zero-error for every model cannot exploit this rarity, since each stop decision must be made before the text that would settle it exists (\cref{thm:antichain}).
The stop entropy takes \res{all}{e0}{any}{stopmin}--\res{all}{e0}{any}{stopmax}\% of $\Hstr$ in every setting (\cref{fig:decomp}).

\paragraph{SyncPool's collapse comes from a handful of hub tokens.}
SyncPool's pools are the trees of this forest, so its coder sees only the masses $M(r)$ and forgoes all of $\Hbranch$.
A few short \emph{hub} tokens that prefix thousands of others make the trees large.
In Llama-3.1-8B, most word tokens begin with a space, and once the bare space token enters the candidate set, its tree takes in every space-prefixed word.
From $k=16$ to the full vocabulary, the largest tree at a step holds on average \res{llama}{e0}{16}{largest}\% to \res{llama}{e0}{Full}{largest}\% of the probability, and SyncPool's share of $\Hstr$ falls from \res{llama}{e0}{16}{root}\% to \res{llama}{e0}{Full}{root}\%.
In Qwen3-8B-Base, common characters are three-byte strings that are not prefixes of one another, so up to $k=1024$ SyncPool keeps \res{qwen}{e0}{1k}{root}--\res{qwen}{e0}{16}{root}\% of $\Hstr$.
With the full vocabulary, single-byte tokens for UTF-8 lead bytes such as \texttt{0xE5} and \texttt{0xE6} enter the candidate set, and each merges every character that starts with it; SyncPool's share falls to \res{qwen}{e0}{Full}{root}\%.
In llm-jp-3-7.2b, the space and the digits are the hubs at small $k$; from $k=1024$ on, byte-fallback tokens for the lead bytes \texttt{0xE3} (shared by all kana) and \texttt{0xE5} join them, and \texttt{0xE3} alone produces \res{llmjp}{e0}{Full}{topbranchone}\% of $\Hbranch$ at the full vocabulary.
SyncPool's share falls gradually, from \res{llmjp}{e0}{16}{root}\% to \res{llmjp}{e0}{Full}{root}\%.

\paragraph{Byte and whitespace hubs inflate $\Hbranch$; lexical prefixes set $\Hstop$.}
Most stop decisions are nearly certain.
Weighted by the probability $M(v)$ of reaching them, \res{llama}{e0}{Full}{polar}\%, \res{qwen}{e0}{Full}{polar}\% and \res{llmjp}{e0}{Full}{polar}\% of nodes on Llama-3.1-8B, Qwen3-8B-Base and llm-jp-3-7.2b have $q_v<0.1$ or $q_v\ge0.9$ at the full vocabulary.
Because the model almost never stops mid-character, prefixes of multi-byte characters sit near $q_v=0$.
At the full vocabulary, the top single-byte UTF-8 lead tokens produce \res{qwen}{e0}{Full}{leadbranch}\% and \res{llmjp}{e0}{Full}{leadbranch}\% of $\Hbranch$ on Chinese and Japanese, yet none of them ranks among the top ten contributors to $\Hstop$.
The bare space on English, whose tree has mean mass $M(\text{\texttt{\textvisiblespace}})\approx\res{llama}{e0}{Full}{spacemass}$, likewise produces \res{llama}{e0}{Full}{spacebranch}\% of $\Hbranch$ but only \res{llama}{e0}{Full}{topstopone}\% of $\Hstop$.
The rest of $\Hstop$ is spread over complete forms that can either end a token or take a continuation, with at most \res{all}{e0}{Full}{topstopmax}\% from any single node: function words, stems and punctuation on English (\texttt{\textvisiblespace a}, \texttt{\textvisiblespace the}, \texttt{.}), single-character morphemes and punctuation on Chinese (\emph{de}, \emph{shi}, \emph{zhong}), and case or topic particles on Japanese (\emph{no}, \emph{ni}, \emph{to}, \emph{de}, \emph{o}, \emph{wa}, \emph{ga}) that also begin compound particles.
Hence $\Hbranch$ rises from \res{all}{e0}{16}{branchmin}--\res{all}{e0}{16}{branchmax}\% of $\Hstr$ at $k=16$ to \res{all}{e0}{Full}{branchmin}--\res{all}{e0}{Full}{branchmax}\% at the full vocabulary, while $\Hstop$ stays level.

\paragraph{Dropping hubs recovers capacity only by giving up exactness.}
Hub-drop, which deletes the nodes with $q_v<0.05$ before running SyncPool, leaves the coder \res{llama}{e0}{Full}{hubdrop}\%, \res{qwen}{e0}{Full}{hubdrop}\% and \res{llmjp}{e0}{Full}{hubdrop}\% of $\Hstr$ at the full vocabulary.
For every model, $k$ and threshold in $\{0.01,0.05,0.1,0.2\}$, it stays below coin projection's budget and is not exact.

\section{Coin Projection}
\label{sec:method}

\subsection{Construction}
\label{sec:coin}

\begin{figure*}[t]
\centering
\includegraphics[width=\textwidth]{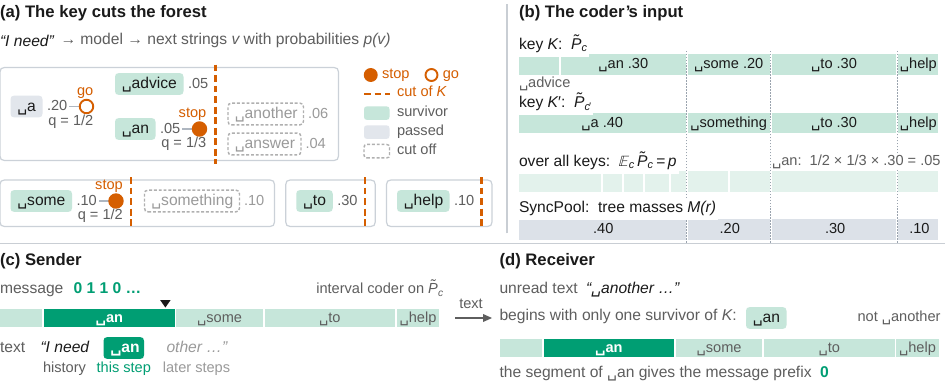}
\caption{The key decides every stop, so both ends stay on one token history; probabilities are illustrative.
(a)~The coin of key $K$ at node $v$ stops with probability $q_v=p(v)/M(v)$; the survivors, where the walk stops, are prefix-free.
(b)~Under every key each tree keeps its mass, and $\E_c\Pt_c=p$ (\cref{prop:coin}).
(c,\,d)~The message picks a survivor through the coder (an interval coder here); the receiver takes the only survivor of $K$ that begins the unread text, \texttt{\textvisiblespace an}, although retokenization would read \texttt{\textvisiblespace another}, and both ends append it to the history.}
\label{fig:method}
\end{figure*}

Coin projection fixes every stop/go decision in advance with shared randomness and leaves the coder only the choices of root and child (\cref{fig:method}).
Under a fixed key, the walk never passes a node whose coin says stop, so no string it can end at is a prefix of another.
Each coin stops with the walk's own probability $q_v$, so over keys every string keeps its probability.

\begin{definition}[Coin projection]
\label{def:coin}
Let $c=(c_v)_{v\ \mathrm{node}}$ with $c_v\in\{0,1\}$ be stop coins.
A string $v\in S$ \emph{survives} if $c_a=0$ for every node $a\pfx v$, and $v$ is a leaf or $c_v=1$.
The \emph{survivor law} is
\begin{equation}
\label{eq:survivor}
\Pt_c(v)=M(v)\prod_{a\in S,\ a\pfx v}\frac{M(a)}{M(a)-p(a)}
\end{equation}
if $v$ survives, and $\Pt_c(v)=0$ otherwise.
When the coins are independent with $\Pr[c_v=1]=q_v$, $\Pt_c$ is the \emph{coin projection} of $p$.
\end{definition}

$\Pt_c$ is the law of the walk above once the coins have decided where to stop.
The walk then stops at $v$ if $c_v=1$ and otherwise moves to a child.
The denominators are positive because a node has a child of positive probability.

\begin{proposition}[Coin projection is exact and spends all but the stop entropy]
\label{prop:coin}
For every coin vector $c$:
\begin{enumerate}[label=(\alph*),nosep,leftmargin=*]
\item the survivors form a prefix antichain;
\item for every root $r$, the survivors in the tree of $r$ carry total probability $M(r)$ under $\Pt_c$.
\end{enumerate}
Over independent coins:
\begin{enumerate}[label=(\alph*),nosep,leftmargin=*,resume]
\item $\E_c\,\Pt_c=p$;
\item $\E_c\,H(\Pt_c)=H(p)-\Hstop$.
\end{enumerate}
\end{proposition}

Part (a) holds because a surviving node stopped, so none of its extensions survives, and (b) because the walk entering root $r$ ends at a survivor of that tree.
For (c), $v$ survives with probability $\prod_{a\pfx v}(1-q_a)\cdot q_v$ (with $q_v=1$ for a leaf), which cancels the product in \cref{eq:survivor}.
For (d), given $c$, the walk randomizes only over roots and over children at the nodes it passes, and it passes $v$ with probability $M(v)-p(v)$ on average (\cref{app:proofs}).

\paragraph{Parsing.}
The unread text starts with the sender's string $v$.
Any other survivor that is also a prefix of the unread text would be comparable with $v$, which part (a) rules out.
The receiver therefore finds the string without lookahead, and it calls the model only on the true token history.
\Cref{alg:coin} in \cref{app:alg} gives both ends as pseudocode.

\paragraph{Relation to SyncPool.}
By part (b), merging each tree's survivors turns $\Pt_c$ into the tree masses, which are SyncPool's coder input (\cref{sec:cost}).

\begin{corollary}[Coin refines SyncPool for every coin vector]
\label{cor:refine}
For every coin vector $c$, SyncPool's coder input is a coarsening of $\Pt_c$.
Hence $H(\Pt_c)\ge\Hroot$ for every $c$, and $\E_c H(\Pt_c)-\Hroot=\Hbranch$.
\end{corollary}

\subsection{Optimality in the single-history class}
\label{sec:optimal}

\begin{definition}[Single-history exact schemes]
\label{def:class}
Sender and receiver share randomness $K$ (the key), a prompt and a public truncation rule.
The message $m$ is uniform and independent of $K$.
For each step $t$, let the \emph{coder input} $P^K_t(\cdot\mid x_{<t})$ be the conditional law of the token $X_t$ given $K$ and the history $X_{<t}=x_{<t}$.
We model the channel as a stream: once the message is embedded, the sender keeps sampling, and the receiver may read as far ahead as it likes.
A scheme belongs to the class if it has these three properties for every full-support language model.
\begin{enumerate}[label=(\roman*),nosep,leftmargin=*]
\item \emph{Exact}: the token sequence has the law of ordinary sampling under the truncation rule.
\item \emph{Single history}: $P^K_t$ depends on the model only via its outputs on prefixes of $x_{<t}$.
\item \emph{Zero error}: the receiver recovers the token sequence with probability one.
\end{enumerate}
\end{definition}

Condition (ii) is non-look-ahead channel access \citep{hopper2002provably,berndt2018gold} with next-token laws in place of samples, and requiring (i) and (iii) across models makes the scheme universal \citep{dedic2009upper}.
A scheme tailored to one known model could build in every output and give (ii) no force; for a deployed model, off-history outputs vary with the context, so beating the bound requires extra forward passes (\cref{app:proofs}) or grey-box assumptions \citep{liskiewicz2013grey}.
The receiver may call the model on any sequence.
SyncPool and coin projection are in the class; Basic, MWIS, Stepwise and hub-drop violate (i), multi-history designs (ii), and Native and ReTokSync (iii).

\begin{theorem}[A single history forces antichains]
\label{thm:antichain}
Assume the vocabulary has a token for every byte, as byte-level BPE and SentencePiece with byte fallback do.
In every scheme of the class, for almost every key and every history of positive probability, the support of $P^K_t$ has at most one token per string, and its strings form a prefix antichain.
\end{theorem}

The proof in \cref{app:proofs} constructs a full-support model under which the shorter candidate is likely continued into the longer one and both then into the same text, so no receiver can separate them.
Lookahead decodes some codes that are not prefix-free, such as $\{0,01\}$, where no codeword starts with 1, but a full-support model can continue a string with any bytes, so lookahead does not help.

\begin{theorem}[The stop-entropy bound]
\label{thm:bound}
Let $\kappa$ be a random variable, and let $(P_\kappa)$ be string laws with $\E_\kappa P_\kappa=p$ such that every $\supp P_\kappa$ is a prefix antichain.
If $X\mid\kappa\sim P_\kappa$, then $I(X;\kappa)\ge\Hstop$, or equivalently
\begin{equation}
\label{eq:bound}
\E_\kappa H(P_\kappa)\ \le\ H(p)-\Hstop.
\end{equation}
Equality holds if and only if the following laws do not depend on $\kappa$: the law of the root, and, at every node, the law of the child given that $X$ passes the node.
In that case, every $P_\kappa$ is a survivor law \eqref{eq:survivor}, and along every root-to-leaf chain the first stop has the same law as under independent coins.
In other words, $P_\kappa$ is the coin projection up to how the stop decisions of incomparable nodes are coupled.
\end{theorem}

\begin{proof}[Proof sketch]
Write $X$ as a sequence of decisions along the forest: the root, then stop or go at each node reached, then the child.
The chain rule expands $I(X;\kappa)$ into one nonnegative term per decision.
Given $\kappa$, the stop/go decision at a reached node $v$ is deterministic.
If $v\in\supp P_\kappa$, the antichain property excludes every extension of $v$, so the walk stops; otherwise it goes.
This decision is reached with probability $M(v)$ and has marginal entropy $\hb(q_v)$, so its term equals $M(v)\,\hb(q_v)$.
Summing over nodes gives $\Hstop$.
Root and child terms vanish exactly when their laws ignore $\kappa$.
\end{proof}

\Cref{thm:antichain} confines every key to independent sets of the prefix comparability graph, so the bound is its graph entropy \citep{korner1973coding,alon1996source}; the value also follows from classical rules \citep{simonyi1995graph}, and \cref{thm:bound} characterizes every optimal mixture.

\begin{corollary}[Budget of single-history schemes]
\label{cor:budget}
For every scheme of the class and every $T$, the first $T$ tokens carry at most
\[
I(m;X_{1:T}\mid K)\le\textstyle\sum_{t\le T} \E\bigl[\Hstr(p_t)-\Hstop(p_t)\bigr]
\]
bits about the message, where $p_t$ is the string law at step $t$.
Even if the receiver needs only the message, any $m$ decoded without error has $H(m)=I(m;X_{1:T}\mid K)$ and obeys the same bound; coin projection meets every per-step term with equality (\cref{app:proofs}).
\end{corollary}

\subsection{Implementation}
\label{sec:impl}

\paragraph{Static prefix table.}
For each tokenizer we precompute all proper byte-prefix pairs after merging homographs (\cref{tab:models}).
Each step masks this table to the top-$k$ set and sums masses with scatter-adds, so no tree is rebuilt (\cref{app:details}).

\paragraph{Determinism and coders.}
The projection runs on the CPU in \texttt{float64} in a fixed reduction order, so both ends compute the same $\Pt$ bit for bit.
It only changes the distribution handed to the coder, so the coders of \cref{sec:prelim} accept $\Pt$ unchanged.
Since disambiguation alone consumes at least $\Hstop$ bits of shared randomness per step (\cref{thm:bound}), we expand a finite key through a pseudo-random function (PRF) with a fresh nonce and separate labels, so exactness holds computationally (\cref{prop:security}).

\section{Experiments}
\label{sec:experiments}

\begin{table*}[t]
\centering
\footnotesize
\setlength{\tabcolsep}{2.6pt}
\renewcommand{\arraystretch}{0.92}
\begin{tabular}{@{}ll ccc ccc ccc ccc@{}}
\toprule
 &  & \multicolumn{3}{c}{$k=16$} & \multicolumn{3}{c}{$k=256$} & \multicolumn{3}{c}{$k=4096$} & \multicolumn{3}{c}{Full vocabulary} \\
\cmidrule(lr){3-5}\cmidrule(lr){6-8}\cmidrule(lr){9-11}\cmidrule(lr){12-14}
Model & Method & BPT$\uparrow$ & Util$\uparrow$ & $\KL_{\mathrm{sec}}\downarrow$ & BPT$\uparrow$ & Util$\uparrow$ & $\KL_{\mathrm{sec}}\downarrow$ & BPT$\uparrow$ & Util$\uparrow$ & $\KL_{\mathrm{sec}}\downarrow$ & BPT$\uparrow$ & Util$\uparrow$ & $\KL_{\mathrm{sec}}\downarrow$ \\
\midrule
 \multirow{8}{*}{\shortstack[l]{Llama-3.1-8B\\(EN, 128k)}} & Native$^\ast$ & 2.02 & .644 & 0 & 3.17 & .890 & 0 & 3.61 & .971 & 0 & 3.79 & .985 & 0 \\
  & Basic & 2.25 & .635 & .466 & 4.35 & .916 & .933 & 5.55 & .984 & 1.02 & --- & --- & --- \\
  & MWIS & 1.90 & .602 & .052 & 2.81 & .824 & .074 & 3.34 & .901 & .088 & 3.24 & .897 & .088 \\
  & Stepwise & 2.03 & .645 & .0019 & 3.19 & .891 & .002 & 3.69 & .976 & .0043 & --- & --- & --- \\
  & Hub-drop & 1.79 & .567 & .0019 & 2.70 & .764 & .0046 & 3.04 & .810 & .0066 & 3.20 & .832 & .0073 \\
\cmidrule(lr){2-14}
  & SyncPool & 1.61 & .511 & 0 & 0.85 & .241 & 0 & 0.49 & .131 & 0 & 0.52 & .133 & 0 \\
  & LAS & 1.98 & .631 & 0 & 3.12 & .885 & 0 & 3.71 & .979 & 0 & 3.79 & .979 & 0 \\
  & \textbf{Coin (ours)} & 1.85 & \textbf{.592} & 0 & 2.86 & \textbf{.810} & 0 & 3.35 & \textbf{.891} & 0 & 3.52 & \textbf{.910} & 0 \\
\midrule
 \multirow{8}{*}{\shortstack[l]{Qwen3-8B-Base\\(ZH, 152k)}} & Native$^\ast$ & 1.73 & .628 & 0 & 2.85 & .857 & 0 & 3.76 & .969 & 0 & 4.05 & .992 & 0 \\
  & Basic & 1.95 & .616 & .612 & 3.98 & .903 & .841 & 4.70 & .951 & .811 & --- & --- & --- \\
  & MWIS & 1.47 & .563 & .049 & 2.49 & .790 & .070 & 3.07 & .879 & .086 & 3.23 & .892 & .091 \\
  & Stepwise & 1.74 & .613 & .026 & 2.94 & .857 & .029 & 3.97 & .973 & .022 & --- & --- & --- \\
  & Hub-drop & 1.44 & .540 & .0011 & 2.52 & .749 & .0021 & 3.26 & .840 & .0028 & 3.25 & .844 & .0029 \\
\cmidrule(lr){2-14}
  & SyncPool & 1.47 & .552 & 0 & 2.55 & .734 & 0 & 2.99 & .767 & 0 & 1.63 & .423 & 0 \\
  & LAS & 1.67 & .620 & 0 & 2.85 & .850 & 0 & 3.79 & .967 & 0 & 4.02 & .983 & 0 \\
  & \textbf{Coin (ours)} & 1.58 & \textbf{.577} & 0 & 2.68 & \textbf{.801} & 0 & 3.40 & \textbf{.891} & 0 & 3.85 & \textbf{.923} & 0 \\
\midrule
 \multirow{8}{*}{\shortstack[l]{llm-jp-3-7.2b\\(JA, 100k)}} & Native$^\ast$ & 1.72 & .569 & 0 & 2.85 & .830 & 0 & 3.77 & .956 & 0 & 4.03 & .989 & 0 \\
  & Basic & 1.81 & .560 & .456 & 3.53 & .852 & .689 & 4.39 & .938 & .708 & --- & --- & --- \\
  & MWIS & 1.45 & .518 & .045 & 2.52 & .772 & .067 & 3.15 & .867 & .085 & 3.37 & .873 & .097 \\
  & Stepwise & 1.65 & .561 & .0029 & 2.83 & .823 & .0027 & 3.65 & .963 & .003 & --- & --- & --- \\
  & Hub-drop & 1.35 & .499 & .0011 & 2.47 & .720 & .0027 & 3.21 & .795 & .0043 & 3.55 & .832 & .0046 \\
\cmidrule(lr){2-14}
  & SyncPool & 1.39 & .495 & 0 & 2.26 & .661 & 0 & 2.13 & .552 & 0 & 1.40 & .336 & 0 \\
  & LAS & 1.69 & .572 & 0 & 2.87 & .834 & 0 & 3.90 & .955 & 0 & 4.01 & .984 & 0 \\
  & \textbf{Coin (ours)} & 1.52 & \textbf{.514} & 0 & 2.76 & \textbf{.774} & 0 & 3.52 & \textbf{.876} & 0 & 3.90 & \textbf{.920} & 0 \\
\bottomrule
\end{tabular}

\caption{Coin projection is the only evaluated single-history method that is exact, zero-error and free of SyncPool's collapse; among exact, zero-error methods, only LAS codes more.
$\KL_{\mathrm{sec}}$ in nats per step.
Bold: Util whose 95\% paired interval of the ratio to SyncPool's excludes one. $^\ast$Native lost \res{llama}{native}{all}{err}, \res{qwen}{native}{all}{err} and \res{llmjp}{native}{all}{err} of \res{all}{native}{any}{n} texts (EN, ZH, JA); no other decoded text failed. Basic and Stepwise run up to $k=4096$, as in the official code (\cref{app:details}).
}
\label{tab:main}
\end{table*}

\subsection{Setup}
\label{sec:setup}

\paragraph{Models and data.}
We evaluate one base model per language on C4 continuations \citep{raffel2020exploring,xue2021mt5}: Llama-3.1-8B on English (128k byte-BPE; \citealp{grattafiori2024llama}), Qwen3-8B-Base on Chinese (152k byte-BPE; \citealp{yang2025qwen3}), and llm-jp-3-7.2b on Japanese (100k byte-fallback SentencePiece; \citealp{llmjp2024llmjp}).
Contexts are the first 10 English words or 10 Chinese or Japanese characters \citep{yan2025addressing}.

\paragraph{Protocol.}
Every text has $T=128$ tokens, plus at most 8 to end on a UTF-8 character boundary, sampled at temperature 1 with $k\in\{16,64,256,1024,4096\}$ or the full vocabulary.
Special tokens, including the end of text, are masked, so a stegotext follows the length rule of its cover.
Each text embeds a uniformly random bit stream, and its payload is the longest stream prefix that the coder's choices fix.
Contexts, keys and streams are paired across methods and $k$, with \res{all}{e1}{any}{nmin} texts per cell (results for $k=64$ and $1024$ in \cref{app:details}).
All methods use the exact range coder of \citet{yan2026efficient} and their own receivers, which get only the Unicode text.
MWIS is solved exactly on the prefix forest, LAS \citep{feng2025high} runs from its public release with its threshold $\tau=0$, and the other filters and SyncPool use the official code on byte strings (\cref{app:details}).
We decode all texts of Native, SyncPool and coin projection and the first \res{all}{e1}{any}{recvmin} per cell of the others.

\paragraph{Metrics.}
We report the \emph{extraction failure rate} (texts whose decoded bits differ from the payload), \emph{bits per token} (BPT), \emph{utilization} (payload bits over the full-model entropy along the method's own text, so losses to truncation and the coder count against it), and the \emph{security divergence} $\KL_{\mathrm{sec}}$ (per-step KL of the key-averaged coder law from top-$k$ sampling, zero by construction for exact schemes).

\subsection{Main results}
\label{sec:main}

\paragraph{Coin projection removes SyncPool's collapse.}
Both are exact and lost no text (\cref{tab:main}), but at the full vocabulary SyncPool's utilization fell to \res{llama}{syncpool}{Full}{utilpct}\%, \res{qwen}{syncpool}{Full}{utilpct}\% and \res{llmjp}{syncpool}{Full}{utilpct}\% on English, Chinese and Japanese, while coin projection kept \res{llama}{coin}{Full}{utilpct}\%, \res{qwen}{coin}{Full}{utilpct}\% and \res{llmjp}{coin}{Full}{utilpct}\%.
At their best $k$, coin projection's BPT was \res{all}{bestk}{syncpool}{min}--\res{all}{bestk}{syncpool}{max} times SyncPool's (\cref{tab:bestk}).

\paragraph{LAS buys the stop entropy with model calls.}
LAS came close to Native's utilization at every $k$ and lost no decoded text; at the full vocabulary it exceeded coin projection's by \res{all}{las}{Full}{utilgapmin}--\res{all}{las}{Full}{utilgapmax} points, most of the share that $\Hstop$ takes.
It paid with \res{qwen}{time@las}{16}{fwdpertok}--\res{llama}{time@las}{16}{fwdpertok} forward passes per token at $k=16$ and \res{qwen}{time@las}{Full}{fwdpertok}--\res{llama}{time@las}{Full}{fwdpertok} at the full vocabulary, and its unbatched sender took \res{all}{time@lascoin}{Full}{min}--\res{all}{time@lascoin}{Full}{max} times coin projection's time (\cref{app:las}).

\paragraph{What remains is the stop entropy.}
Across models and $k$, coin projection's utilization reached a median \res{all}{ratio}{native}{median}\% of Native's and \res{all}{ratio}{stepwise}{median}\% of Stepwise's, which hand the coder almost the whole law.
This matches the \res{all}{e0}{any}{coinmin}--\res{all}{e0}{any}{coinmax}\% of $\Hstr$ left after $\Hstop$ (\cref{sec:cost}), the price of exactness and zero error under a single history (\cref{thm:bound}).
Native pays with failures, most of them silent (\res{llama}{native}{all}{silent}, \res{qwen}{native}{all}{silent} and \res{llmjp}{native}{all}{silent} of its lost texts decoded to wrong bits; \cref{app:native-err}); regenerating failed texts shifts the law by up to \res{all}{rej}{128}{klmax} nats per text (\cref{app:rejection}).

\paragraph{Stepwise buys the stop entropy with divergence.}
Stepwise keeps a string and its extensions, so the message decides where the token stops.
It then rejects any next token that would merge with the last one, so a step lies $-\log(1-r)$ nats from top-$k$ sampling for the rejected mass $r$.
Over a text, this exceeds the shift of regenerating Native's failed texts, and on Chinese, rejecting tokens ending inside a character raises it about tenfold (\cref{app:details}).

\paragraph{ReTokSync repairs errors instead.}
ReTokSync resets its coder whenever retokenization changes the tokens and later repairs its wrong bits.
The reset needs a coder that emits bits step by step, so we run it on Discop (\cref{app:retoksync}).
Its utilization was \res{all}{rts}{any}{rtscoinmin}--\res{all}{rts}{any}{rtscoinmax} times coin projection's, with wrong bits in \res{all}{rts}{any}{texterrmin}--\res{all}{rts}{any}{texterrmax}\% of texts before repair.
It is exact against a cover that retokenizes the text before every step, not ordinary sampling; \cref{thm:bound} holds there too.

\paragraph{Filters trade security for capacity.}
Basic moves the mass of every internal node to the leaves and can exceed coin projection's utilization, at up to \res{all}{basic}{any}{klmax} nats per step.
MWIS, which keeps a heaviest prefix-free set, has the smallest divergence of any key-independent prefix-free filter (\cref{rem:filter}) but is not exact.
A $\chi^2$ next-token test rejected Native, SyncPool and coin projection in \res{all}{gof}{exact}{rejtot} of \res{all}{gof}{exact}{ctxtot} tests, as chance predicts, and MWIS and Basic on at least \res{all}{gof}{ctrl}{rejmin} of 20 histories per cell (\cref{tab:gof}).

\subsection{Gain prediction and runtime}
\label{sec:predict}

\begin{figure}[t]
\centering
\includegraphics[width=\columnwidth]{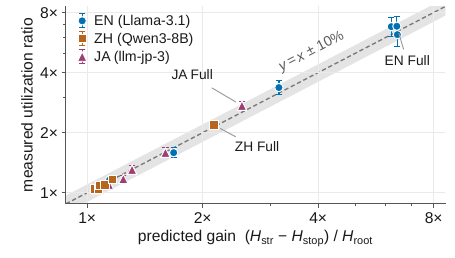}
\caption{The forward-only decomposition predicts the end-to-end gain of coin projection over SyncPool.
Each point is one model and $k$: $x$ is $(\Hstr-\Hstop)/\Hroot$, and $y$ is the ratio of the two utilizations in \cref{tab:main} with its 95\% interval; dashed, $y=x$; shaded, $\pm10\%$.}
\label{fig:predict}
\end{figure}

By \cref{prop:coin}(d) and \cref{cor:refine}, the coder budgets of coin projection and SyncPool stand in the ratio $(\Hstr-\Hstop)/\Hroot=1+\Hbranch/\Hroot$, which needs only forward passes of ordinary sampling.
\Cref{fig:predict} compares this ratio with the measured ratio of utilizations, and every point lies within 10\% of the diagonal over a sixfold range.
The predicted gain rises with $k$ to \res{llama}{gain}{Full}{pred} on English and \res{llmjp}{gain}{Full}{pred} on Japanese; on Chinese, where characters form separate roots, it stays at or below \res{qwen}{gain}{4k}{pred} up to $k=4096$ and jumps to \res{qwen}{gain}{Full}{pred} once single-byte UTF-8 prefixes enter the candidate set.
In \res{all}{gain}{ci}{inside} of \res{all}{gain}{cells}{n} cells, the prediction lies inside the 95\% interval of the measured gain, off by \res{all}{gain}{dev}{total}\% on average; evaluated on coin projection's own texts, it does so in \res{all}{gain}{citraj}{inside} (\cref{app:predict}).
The gain and its prediction carry over to 512-token texts, temperature and nucleus sampling, and longer prompts, except that the gain exceeds the prediction where SyncPool's texts carry only a few dozen bits (\cref{app:robust}); the gain also carries over to other coders (\cref{app:coders}).

\paragraph{Computational cost.}
Timed on one RTX 5090 (\cref{app:timing}), coin projection added at most \res{llama}{time@coin}{4k}{sel}\,ms per token to the \res{llama}{time@coin}{4k}{fwd}\,ms forward pass on English at $k\le 4096$.
At the full vocabulary, its projection took \res{llama}{time@coin}{Full}{sel}\,ms, against \res{llama}{time@syncpool}{Full}{sel}\,ms for SyncPool's official grouping and \res{llama}{time@mwis}{Full}{sel}\,ms for exact MWIS; Stepwise took \res{llama}{time@stepwise}{4k}{sel}\,ms already at $k=4096$ (\cref{tab:timing}).
At each method's highest-throughput $k$, coin projection sent 1.20--1.87 times SyncPool's bits per second and 0.87--0.89 times Native's.

\section{Related Work}
\label{sec:related}

\paragraph{Tokenization ambiguity in steganography.}
Beyond the designs of \cref{sec:prelim}, \citet{yan2024near} verify candidates against the history, \citet{yan2025addressing} measure how often retokenization breaks extraction (see also \citealp{bauer2024leveraging}) and note that a theoretical account is lacking, and \citet{yan2026comprehensive} list high capacity under provable security as an open challenge.
TI-StegoAlign \citep{zhou2026tistegoalign} post-trains the model against inconsistent outputs, changing the base law; STEAD \citep{qi2025stead} embeds with a diffusion language model and error-correcting codes.
Alkaid \citep{cao2026alkaid} and ASW \citep{yan2026anchored} target robustness to edits; Alkaid carries $\approx 0.2$ bits per token and risks errors that edit-free channels avoid.

\paragraph{From tokens to bytes.}
Prefix masses $M(v)$ and stop probabilities $q_v$ appear when token-level models are converted to byte or character laws \citep{loula2025syntactic,vieira2025language,phan2025exact,hayase2026sampling} or likelihoods are marginalized over tokenizations \citep{cao2021you,chirkova2023should}; these methods evaluate the model on several tokenizations, whereas coin projection fixes the stop decisions with shared randomness on one history.
Non-canonical tokenizations affect evaluation and pricing \citep{geh2024where,chatzi2025tokenization}; canonical models exclude them \citep{vieira2025canonical}.

\section{Conclusion}
\label{sec:conclusion}

Every universal single-history scheme that is exact and zero-error must pay the stop entropy for missing token boundaries, however rare ambiguous text is; coin projection attains this bound and returns to the coder the hub trees that SyncPool gives up.
It keeps \res{all}{e0}{any}{coinmin}--\res{all}{e0}{any}{coinmax}\% of $\Hstr$ and reaches \res{all}{gain}{Full}{min}--\res{all}{gain}{Full}{max}$\times$ SyncPool's full-vocabulary utilization with zero divergence and error, within \res{all}{las}{Full}{utilgapmin}--\res{all}{las}{Full}{utilgapmax} points of LAS at one call per token.
How much of $\Hstop$ a few off-history calls or grey-box assumptions recover is open.

\section*{Limitations}

\paragraph{Scope of the bound.}
Our bound covers universal schemes whose sender queries the model only along its emitted token history and that are exact and zero-error across full-support language models (\cref{app:proofs}).
It places no limit on the receiver, and it still holds when the receiver must recover only the message rather than the tokens.
Its stream model sets aside where a finite text ends; under our fixed-length rule, the end can resolve a stop decision between a string $v$ and its extension $v\cdot s$ only at steps after $T-2|s|-3$ (\cref{app:proofs}), and none of the compared schemes uses the end.

\paragraph{Grey-box and hybrid designs.}
The bound is the price of universality across full-support models; a grey-box scheme \citep{liskiewicz2013grey} that needs exactness only for models whose non-canonical continuations are rare can pay less than $\Hstop$ (\cref{app:proofs}).
In the extreme, a model that places no mass on non-canonical tokenizations \citep{vieira2025canonical} makes canonical retokenization exact; making a deployed model canonical, however, changes the law that the stegotext must match.
By contrast, relaxing token-level exactness to text-level exactness across full-support models does not raise the single-history budget when the sender calls the model only along its own history (\cref{app:proofs}); exceeding $\Hstr-\Hstop$ via text-level exactness requires the sender to query alternative tokenizations of the text.
Hybrids that query other tokenizations only at some steps, for example after hub tokens, lie outside the class, and we do not characterize their trade-off between model calls and capacity.

\paragraph{Capacity.}
Coin projection still pays $\Hstop$, \res{all}{e0}{any}{stopmin}--\res{all}{e0}{any}{stopmax}\% of the string entropy in our measurements.
The share is a property of these tokenizers and models, not of the bound.
For a candidate set that is a single chain of nested strings, $\Hstop=\Hstr$ and no exact, zero-error single-history scheme carries anything.
Each compared scheme that codes more than coin projection uses multiple histories, incurs divergence, holds for fewer models, or makes errors (\cref{sec:main}).

\paragraph{Computation.}
As in every provably secure scheme, the receiver needs the model.
With the full vocabulary, our CPU projection and coder add \res{llama}{time@coin}{Full}{sel}\,ms per token on English to the \res{llama}{time@coin}{Full}{fwd}\,ms forward pass of Llama-3.1-8B on one GPU (\cref{tab:timing}); this is less than SyncPool's official implementation needs but more than the forward pass itself.
As with every scheme that codes the model's probabilities, bit-exact agreement between sender and receiver relies on a fixed computation order; STEALTH \citep{fleshman2026stealth} instead tolerates differing probabilities, and even different tokenizers, by absorbing the resulting token flips with an error-correcting code.

\paragraph{Finite precision.}
Exactness here is a property of the law handed to the coder.
Our range coder works with \texttt{float64} cumulative probabilities and the coins with 53-bit uniforms, so the emitted law is exact only up to this rounding.
With the PRF modeled as a random function, the coins move the law of a text's tokens by less than $10^{-11}$ in total variation from that under exact coins, and the range coder moves the law of each token, given the history, by less than $6\cdot10^{-11}$ from the law handed to it (\cref{app:precision}).
\citet{cao2026breaking} detect arithmetic-coding and Meteor stegotext from the low-probability tokens that coarse rounding removes; here the tokens that the coder cannot emit carry less than $1.2\cdot10^{-10}$ of the mass at a step, so ordinary sampling emits such a token less than once in $8\cdot10^{9}$ tokens on average.
This concerns the coder rather than disambiguation, applies equally to every method we compare, and coin projection can feed any coder that closes this gap.

\paragraph{Text length and encoding.}
Our texts follow the fixed length rule of \cref{sec:setup}, and we do not analyze schemes that stop once the message is complete.
Zero error is proven for the byte stream; a Unicode channel carries only valid UTF-8, so over it zero error holds for the texts whose bytes are valid, an event that the sender observes and that has the same probability as under ordinary sampling (\cref{app:details}).

\paragraph{Experimental scale.}
Our models have 7--8B parameters (Llama-3.1-8B, Qwen3-8B-Base and llm-jp-3-7.2b), and we cover three languages and two tokenizer families.
Models beyond 8B in these families share their tokenizers, so the prefix forest and its hub tokens stay the same and only the next-token law changes; the forward-only prediction of \cref{sec:predict} estimates the gain for such a model from its forward passes, but we have not measured extraction end to end beyond 8B.
The receivers of LAS, the filters and hub-drop ran on the first \res{all}{e1}{any}{recvmin} texts of each cell; as with coin projection and SyncPool, their zero error follows from their proofs, and decoding tests our implementations.

\section*{Ethics Statement}
Steganography protects private communication, for example under censorship, and it can also hide harmful content or leaked data from moderation.
Our method does not make stegotext harder to detect than existing provably secure schemes, which already match the model's distribution, though removing extraction failures and SyncPool's capacity loss makes covert misuse over text channels practical as well.
Our experiments use public prompts from C4 and publicly released models, and they involve no human subjects.
We will release code for research use.

\bibliography{refs_base,refs_new}

\appendix
\section{Proofs}
\label{app:proofs}

Throughout, $p$ is a string law with support $S$, and $F$ is its prefix forest (\cref{def:forest}).
For $x\in S$, let $a_0\pfx a_1\pfx\dots\pfx a_m=x$ be the strings of $S$ that are prefixes of $x$, so $a_0$ is a root and $a_{i+1}\in\ch(a_i)$; we call this the \emph{path} of $x$.
The prefixes of $x$ in $S$ form a chain, so the path is unique and $x$ has a unique root.

\paragraph{Decision variables.}
For a random string $X$ with values in $S$, define:
\begin{itemize}[nosep,leftmargin=*]
\item $D_0$, the root of $X$;
\item for every node $v$, the event $A_v=\{v\pfxeq X\}$ that $X$ reaches $v$;
\item on $A_v$, the stop indicator $B_v=\1[X=v]$;
\item on $A_v\cap\{B_v=0\}$, the child $C_v\in\ch(v)$ on the path of $X$.
\end{itemize}
$X$ and its decisions determine each other.
If $X\sim p$, then $\Pr[A_v]=M(v)$, $\Pr[B_v=1\mid A_v]=q_v$, and $\Pr[C_v=c\mid A_v,B_v=0]=M(c)/(M(v)-p(v))$.

\subsection{Lemma 1}
Apply the grouping rule of entropy \citep[\S2]{cover1999elements} to the decision sequence:
\begin{align*}
H(X) &= H(D_0)+\sum_{v}\Pr[A_v]\Bigl(H(B_v\mid A_v)\\
&\quad+\Pr[B_v{=}0\mid A_v]\,H(C_v\mid A_v,B_v{=}0)\Bigr).
\end{align*}
Substituting the three laws above gives $H(p)=\Hroot+\Hstop+\Hbranch$.
For tokens, $X_{\mathrm{tok}}\mapsto(\bytes(X_{\mathrm{tok}}),X_{\mathrm{tok}})$ gives $H(\pi)=H(p)+\sum_v p(v)H(\pi(\cdot\mid\bytes{=}v))=\Hstr+\Hhom$.
\qed

\subsection{Proposition 1}
Fix a coin vector $c$.
Let $W_c$ be the walk that enters root $r$ with probability $M(r)$.
At a node $v$ it stops if $c_v=1$ and otherwise moves to child $c'$ with probability $M(c')/(M(v)-p(v))$; at a leaf it stops.

\emph{(a)} Let $v$ survive and $v\pfx w\in S$.
Then $v$ has an extension in $S$, so it is a node with $c_v=1$.
Since $v\pfx w$ and $c_v=1$, $w$ does not survive.

\emph{(b)} $W_c$ stops at $v$ only if $v$ is a leaf or $c_v=1$, and it passes a node $a$ only if $c_a=0$.
So it always ends at a survivor, and it can reach a survivor only along that survivor's path.
For a survivor $x$ with path $a_0\pfx\dots\pfx a_m=x$, the probability that $W_c$ ends at $x$ telescopes:
\begin{align*}
&M(a_0)\prod_{i<m}\frac{M(a_{i+1})}{M(a_i)-p(a_i)}\\
&\qquad=M(x)\prod_{i<m}\frac{M(a_i)}{M(a_i)-p(a_i)}.
\end{align*}
This is \cref{eq:survivor}, so $\Pt_c$ is the law of $W_c$.
A walk that enters root $r$ stays in the tree of $r$, so the tree of $r$ carries mass $M(r)$.
The masses $M(r)$ sum to one, so $\Pt_c$ is a probability distribution.

\emph{(c)} Under independent coins, $x$ survives with probability $\prod_{i<m}(1-q_{a_i})\cdot q_x$, where $q_x=1$ if $x$ is a leaf.
This equals $\prod_{i<m}\frac{M(a_i)-p(a_i)}{M(a_i)}\cdot\frac{p(x)}{M(x)}$.
Multiplying by \cref{eq:survivor} gives $\E_c\Pt_c(x)=p(x)$.

\emph{(d)} Given $c$, the grouping rule applied to $W_c$ gives
\[
H(\Pt_c)=\Hroot+\sum_{v\ \mathrm{node}}\Pr\nolimits_c[W_c\text{ passes }v]\cdot H_v.
\]
Here $H_v$ is the entropy of the child law at $v$; stop decisions contribute nothing because $c$ fixes them.
By the telescoping in (b), $\Pr_c[W_c\text{ passes }v]$ is the right-hand side of \cref{eq:survivor} times $\1[c_a=0\ \forall a\pfxeq v]$.
Its expectation over the coins is therefore
\begin{align*}
&M(v)\prod_{a\pfx v}\frac{M(a)}{M(a)-p(a)}\cdot\prod_{a\pfxeq v}(1-q_a)\\
&\quad= M(v)-p(v).
\end{align*}
Hence $\E_c H(\Pt_c)=\Hroot+\Hbranch=H(p)-\Hstop$ by \cref{lem:grouping}.
\qed

The proofs of (c) and (d) use the coins only through the probability that the first stop along each path occurs at a given string.
Any coupling that preserves these probabilities gives the same results.

\subsection{Corollary 1}
Two strings of $S$ are comparable only if they share a root, and every string is comparable with its root.
Hence the connected components of the prefix relation on $S$, which are SyncPool's groups, are exactly the trees of $F$.
SyncPool feeds the coder the group masses $(M(r))_r$.
By \cref{prop:coin}(b), this is the image of $\Pt_c$ under the map that sends each survivor to its root.
A deterministic map cannot increase entropy, so $H(\Pt_c)\ge\Hroot$ for every $c$.
Finally, $\E_c H(\Pt_c)-\Hroot=\Hbranch$ by \cref{prop:coin}(d).
\qed

\subsection{Theorem 1}
\paragraph{Model.}
A language model maps every token sequence, prompt included, to a law over the vocabulary; unlike the sampling oracle of \citet{hopper2002provably} and \citet{dedic2009upper}, querying a prefix returns its full next-token law, so a lower bound proved here also holds under sampling access.
The receiver is deterministic given the key $K$, the prompt and the stream $Y$, and it accesses the model as an oracle, on any sequences and in any order; a receiver with randomness of its own is a mixture of such receivers.
For a fixed model, its decisions $\hat X_1,\hat X_2,\dots$ are therefore functions of $K$ and $Y$.

\paragraph{Proof.}
Suppose for contradiction that at some step $t$, history $h$ and distinct tokens $x\ne x'$ with $\bytes(x')=\bytes(x)\cdot s$ ($s$ possibly empty), the event $E=\{X_{<t}=h,\ x,x'\in\supp P^K_t(\cdot\mid h)\}$ has positive probability under a full-support model $\mathrm{LM}_0$.
Then there exist $a,c>0$ and a measurable set $G_0$ of keys with $\mu(G_0)=\Pr[K\in G_0]>0$ such that for every $\kappa\in G_0$, $\Pr[X_{<t}=h\mid K=\kappa]\ge a$ and $P^\kappa_t(x\mid h),P^\kappa_t(x'\mid h)\ge c$.
Fix an infinite byte string $g$ and positive numbers $(\delta_j)_{j\ge 1}$ in $(0,\tfrac12)$ with $\sum_{j=1}^\infty\delta_j\le\frac14 ac\,\mu(G_0)$.
Define a full-support model $\mathrm{LM}^\ast$ that agrees with $\mathrm{LM}_0$ on every sequence extending neither $(\text{prompt},h,x)$ nor $(\text{prompt},h,x')$.
Along the unique single-byte token path spelling $s\cdot g$ after $(\text{prompt},h,x)$, its $j$-th step puts mass $1-\delta_j$ on the $j$-th byte token of $s\cdot g$ and splits $\delta_j$ uniformly over the other $|\Voc|-1$ tokens (and puts uniform weights off that path).
After $(\text{prompt},h,x')$, it does the same for $g$.
Because $x\ne x'$, neither sequence extends the other, so $\mathrm{LM}^\ast$ is well defined and strictly positive everywhere.
Since $1-\delta_j>\tfrac12$, each path token is the most probable token at its step, so every truncation rule that keeps the most probable token (top-$k$, top-$p$, or none) keeps it, and renormalization only raises its mass; the bounds below therefore hold under the truncation rule.

\emph{Step $t$ is unchanged.}
$\mathrm{LM}^\ast$ agrees with $\mathrm{LM}_0$ on every prefix of $(\text{prompt},h)$.
By (ii), the coder inputs up to step $t$ on prefixes of $h$ are identical under $\mathrm{LM}^\ast$, so every $\kappa\in G_0$ still satisfies $\Pr^\ast[X_{<t}=h,X_t=u\mid K=\kappa]\ge ac$ for $u\in\{x,x'\}$.

\emph{Both branches emit the same stream on a common key set.}
Under ordinary sampling from $\mathrm{LM}^\ast$, given $(h,x)$ the continuation fails to follow the single-byte path of $s\cdot g$ with probability at most $\sum_{j=1}^\infty\delta_j\le\frac14 ac\,\mu(G_0)$ by the union bound.
By exactness (i), the unconditional probability of $\{X_{<t}=h,X_t=x,\text{stream}\ne\text{prompt}\cdot h\cdot\bytes(x)\cdot s\cdot g\}$ is at most $\pi^\ast(h,x)\sum_{j\ge 1}\delta_j\le\frac14 ac\,\mu(G_0)$.
On the subset $B_x\subseteq G_0$ of keys $\kappa$ for which the branch $(h,x)$ has conditional probability zero of emitting that target stream, $\{K\in B_x,X_{<t}=h,X_t=x\}$ implies deviating from the target stream, so $ac\,\mu(B_x)\le\frac14 ac\,\mu(G_0)$, giving $\mu(B_x)\le\frac14\mu(G_0)$.
Likewise $\mu(B_{x'})\le\frac14\mu(G_0)$ for the analogous set $B_{x'}\subseteq G_0$ on $(h,x')$.
Hence $G_1=G_0\setminus(B_x\cup B_{x'})$ has $\mu(G_1)\ge\frac12\mu(G_0)>0$, and for every $\kappa\in G_1$ both $\{X_{<t}=h,X_t=x\}$ and $\{X_{<t}=h,X_t=x'\}$ produce the identical infinite byte stream $y^\ast=\text{prompt}\cdot h\cdot\bytes(x')\cdot g$ with positive probability.

\emph{The receiver errs on one branch.}
For $\kappa\in G_1$, both events give the receiver the key $\kappa$ and the stream $y^\ast$ under the same model $\mathrm{LM}^\ast$.
Thus $\hat X_t$ takes one value while $X_t$ takes both $x$ and $x'$ with positive probability, contradicting (iii).
The argument uses nothing about the receiver beyond its inputs. What forces the antichain is that, by (ii), the sender cannot see how $\mathrm{LM}^\ast$ continues the two branches.
\qed

\paragraph{Remarks.}
The sender may constrain later steps however it likes, as long as it calls the model only on prefixes of the emitted history (ii); exactness across full-support models removes that freedom on the two branches.
The bound is therefore the price of universality across models.
A grey-box scheme \citep{liskiewicz2013grey} that needs exactness only for models whose non-canonical continuations are rare can pay less than $\Hstop$ by letting a key keep both $v$ and an extension $w$ and forbid, at the next step, tokens continuing $v$ into $w$, while keys that stop at $v$ compensate for the forbidden mass; on $\mathrm{LM}^\ast$ such compensation fails.

For example, let $w=v\cdot s$ be the only candidate extending $v$, with $s$ a single byte so that any continuation of $v$ into a string starting with $w$ begins at the next token with byte $s$, and let $\eta$ be the mass that the model puts after $v$ on tokens whose strings start with $s$.
Let a fraction $\lambda$ of keys leave the choice between $v$ and $w$ to the message, and let the other keys stop at $v$ with probability $q_v$, as coin projection does.
After $v$, the keys of the first group exclude the tokens starting with $s$, and the keys that stopped at $v$ raise the mass of these tokens to $\eta/(1-\lambda)$, which keeps the next step exact if and only if $\lambda\le 1-\eta$.
Compared with coin projection, the coder gains $\lambda M(v)\hb(q_v)$ bits at $v$ and loses $p(v)\bigl(\hb(\eta)-(1-\lambda)\hb(\eta/(1-\lambda))\bigr)\le p(v)\hb(\eta)$ at the next step, because the strings starting with the single byte $s$ form whole trees of the next step's forest, the two groups differ only in their total mass, and $\Hstop$ is linear in tree masses.
With $p(v)=p(w)=\tfrac12$, $\eta=10^{-3}$ and $\lambda=1-\eta$, the net gain is about $0.99$ bits at a step where the bound allows none.
Under (ii), the sender fixes $\lambda$ before seeing $\eta$, an output on $(\text{prompt},h,v)$ off its history, so a fixed $\lambda>0$ is exact only for models with $\eta\le 1-\lambda$ (under $\mathrm{LM}^\ast$, $\eta$ is close to one).
Conversely, one off-history call to read $\eta$ lets the sender set $\lambda=1-\eta$ and exceed $\Hstr-\Hstop$ while staying exact and zero-error for every full-support model.
Exactness for a single fixed model would likewise give (ii) no force, since a scheme built for that model can build in the model's outputs on every sequence.

\paragraph{Text-level exactness.}
If the sender satisfies (ii), replacing token-level exactness (i) by \emph{text-level exactness} (the byte stream has the law of ordinary sampling for every full-support model) leaves the class unchanged.
Fix a full-support model $\mathrm{LM}_0$ and a length $t$, and write $\sigma(u)$ and $\pi_0(u)$ for the probabilities that the scheme and ordinary sampling under $\mathrm{LM}_0$ emit the token sequence $u$ of length $t$.
Draw independent uniformly random infinite byte strings $g^{(u)}$, one for each such $u$, and let $\mathrm{LM}^\ast$ agree with $\mathrm{LM}_0$ on sequences shorter than $t$ tokens after the prompt, while after each $(\text{prompt},u)$ it puts mass $1-\delta_j$ on the $j$-th byte token of $g^{(u)}$ along its single-byte path, as above.
By (ii), the coder inputs of the first $t$ steps query only sequences shorter than $t$ tokens, so the scheme still emits $u$ with probability $\sigma(u)$ under $\mathrm{LM}^\ast$.
Let $y_u=\text{prompt}\cdot\bytes(u)\cdot g^{(u)}$.
Ordinary sampling under $\mathrm{LM}^\ast$ emits $y_u$ with probability at least $\pi_0(u)\prod_j(1-\delta_j)$, and by text-level exactness so does the scheme.
Given the strings $g^{(w)}$ with $w\ne u$, the scheme's stream law on the branch $X_{\le t}=w$ is fixed and has at most countably many atoms, which the independent uniform $g^{(u)}$ avoids almost surely; since $u$ ranges over finitely many sequences, one realization of the strings avoids them for every $u$ at once.
Then only the branch $u$ emits $y_u$ with positive probability, so $\sigma(u)\ge\pi_0(u)\prod_j(1-\delta_j)$.
Letting $\sum_j\delta_j\to 0$ gives $\sigma(u)\ge\pi_0(u)$, and since both sides sum to one over $u$, $\sigma=\pi_0$.
This holds for every $t$, which is token-level exactness (i); the argument uses only (ii) and text-level exactness.
\paragraph{Recovering only the message.}
Zero error in (iii) asks for the tokens, whereas a receiver may need only the message $m$; this does not raise the budget.
Let $\rho$ collect the sender's randomness beyond $(K,m)$, including the sampling after the message is embedded, so that $\rho$ is independent of $(K,m)$ and, for a fixed model, the tokens are a function of $(m,K,\rho)$.
Suppose that the sender calls the model only on prefixes of its history and that some receiver recovers $m$ from $K$ and the stream with probability one for every full-support model.
Take $(K,\rho)$ as the key.
The coder inputs given $(K,\rho)$ still satisfy (i) and (ii), and a receiver that decodes $m$ and then reruns the sender on $(m,K,\rho)$ recovers every token, so it satisfies (iii).
\Cref{thm:antichain} therefore applies with $(K,\rho)$ in place of $K$.
Since $m$ is independent of $\rho$ given $K$,
\begin{align*}
I(m;X_{1:T}\mid K)&\le I(m;X_{1:T},\rho\mid K)\\
&=I(m;X_{1:T}\mid K,\rho),
\end{align*}
and the proof of \cref{cor:budget} with $(K,\rho)$ in place of $K$ bounds the right side by the same sum.
In particular, a coder that merges prefix-related candidates into one message symbol and lets $\rho$ choose among them cannot exceed the bound, because for almost every $(K,\rho)$, the tokens that the message can select still have prefix-free strings.

\paragraph{Finite texts.}
The stream model sets aside where a text ends, and a receiver that sees the end can separate some stop decisions. In a text of one token, for example, a key may keep both $v$ and $v\cdot s$.
Under our length rule, a text ends at the first step $t'\ge T$ after which its bytes are valid UTF-8 (\cref{app:details}), and the end helps only near $T$.
In the proof above, let $b$ be a string of at most three bytes that completes the last character of $\text{prompt}\cdot h\cdot\bytes(x')$ (if no such string exists, no text through either branch ends), let $z$ be a token that spells two ASCII letters, as in all three of our tokenizers, and let $g$ be an ASCII string.
Let $\mathrm{LM}^\ast$ continue $(\text{prompt},h,x)$ with the single-byte tokens of $s\cdot b$, then $|s|$ copies of $z$, then the single-byte tokens of $g$, and continue $(\text{prompt},h,x')$ with the single-byte tokens of $b\cdot\bytes(z)^{|s|}\cdot g$, with masses $1-\delta_j$ as before.
From step $t+2|s|+|b|$ on, both branches have the same text after the same number of tokens.
If $t+2|s|+|b|\le T$, neither branch can end earlier, so both end at the same step with the same text, the receiver errs on one of them, and the proof goes through.
A key can therefore keep a string and its extension by $s$ only at steps $t>T-2|s|-3$, and at every step it keeps at most one of two homographs, whose branches coincide from step $t$ on.
Since $|s|<128$, the number of steps at which the end can help does not grow with $T$, so the per-token budget of \cref{cor:budget} holds as $T$ grows.

\subsection{Theorem 2}
Let $X\mid\kappa\sim P_\kappa$.
Since $\E_\kappa P_\kappa=p$, every $P_\kappa$ is supported in $S$ (almost surely).
Apply the grouping rule to each $P_\kappa$ and average over $\kappa$.
Conditioning the joint law of $(X,\kappa)$ on $A_v$ reweights $\kappa$ by $P_\kappa(A_v)/M(v)$.
This gives the conditional version of the identity in Lemma 1:
\begin{align*}
H(X\mid\kappa) &= H(D_0\mid\kappa)\\
&\quad+\sum_v M(v)\Bigl(H(B_v\mid\kappa,A_v)\\
&\quad+(1-q_v)\,H(C_v\mid\kappa,A_v,B_v{=}0)\Bigr),
\end{align*}
and subtracting this from Lemma 1 gives
\begin{align}
\label{eq:chain}
I(X;\kappa) &= I(D_0;\kappa)+\sum_v M(v)\Bigl(I(B_v;\kappa\mid A_v)\nonumber\\
&\quad+(1-q_v)\,I(C_v;\kappa\mid A_v,B_v{=}0)\Bigr).
\end{align}
Fix $\kappa$ with $P_\kappa(A_v)>0$.
If $v\in\supp P_\kappa$, the antichain property excludes every proper extension of $v$, so $A_v$ implies $X=v$.
Otherwise $A_v$ implies $X\ne v$.
Either way, $B_v$ is determined by $\kappa$ on $A_v$.
Hence $H(B_v\mid\kappa,A_v)=0$ and $I(B_v;\kappa\mid A_v)=H(B_v\mid A_v)=\hb(q_v)$.
All other terms of \eqref{eq:chain} are nonnegative, so $I(X;\kappa)\ge\Hstop$.
Since $I(X;\kappa)=H(p)-\E_\kappa H(P_\kappa)$, the second form of \eqref{eq:bound} follows.

\paragraph{Equality.}
Equality holds if and only if $I(D_0;\kappa)=0$ and $I(C_v;\kappa\mid A_v,B_v{=}0)=0$ for every node $v$.
In that case, for almost every $\kappa$:
\begin{itemize}[nosep,leftmargin=*]
\item $P_\kappa(D_0=r)=M(r)$;
\item $P_\kappa(C_v=c\mid A_v,B_v{=}0)=M(c)/(M(v)-p(v))$ whenever the conditioning event has positive probability.
\end{itemize}
Take $x\in\supp P_\kappa$ with path $a_0\pfx\dots\pfx a_m=x$.
The strings $a_i$ with $i<m$ are prefixes of $x$, so they lie outside the support and the walk goes at each of them; at $x$ it stops.
Therefore
\[
P_\kappa(x)=M(a_0)\prod_{i<m}\frac{M(a_{i+1})}{M(a_i)-p(a_i)}
\]
which is \cref{eq:survivor}.
So $P_\kappa$ is the survivor law of the stop pattern ``$v$ stops iff $v\in\supp P_\kappa$''.
Taking expectations, $\E_\kappa P_\kappa=p$ holds if and only if, for every $x$,
\begin{align*}
&\Pr\nolimits_\kappa[\text{first stop on path of }x\text{ is at }x]\\
&\quad=\prod_{i<m}(1-q_{a_i})\,q_x,
\end{align*}
the probability under independent coins.
Conversely, every random stop pattern with these first-stop probabilities attains equality.
\qed

\paragraph{Couplings and graph entropy.}
Our implementation draws one independent coin per node via a single stream expansion per step (\cref{app:details}).
Sharing coins between incomparable nodes, for example one uniform per tree depth (at most 85, 85 and 15 internal levels on our English, Chinese and Japanese tokenizers), remains optimal by \cref{thm:bound}.
SyncPool's member draw is an antichain mixture too, but its child choice depends on the shared randomness; it is therefore strictly below the bound whenever $\Hbranch>0$.
The least $I(X;\kappa)$ over variables $\kappa$ whose conditional supports are independent sets of a graph $G$ is \citeauthor{korner1973coding}'s \citeyearpar{korner1973coding} graph entropy $H(G,p)$, which \citet{alon1996source} relate to zero-error source coding.
Here $G$ is the comparability graph of the prefix forest $F$, a trivially perfect graph ($P_4$- and $C_4$-free) whose independent sets are prefix antichains and whose cliques (independent sets of $\bar G$) are chains of $F$.
The value $H(G,p)=\Hstop$ also follows from two classical rules \citep{simonyi1995graph}: the graph entropy of a disjoint union is the average over its parts, weighted by their masses, and a join of two parts adds the binary entropy of the part that contains $X$.
A tree is its root joined to the disjoint union of its subtrees, and unrolling these rules over $F$ gives $\sum_v M(v)\hb(q_v)$.
Drawing a root $r\sim M$ and a child $C_v\sim\gamma_v$ at every node $v$ independently of the stop coins selects a root-to-leaf chain $Z$ intersecting the surviving antichain $Y=\supp\Pt_c$ at $\{X\}$, with $Y\perp Z$, $I(X;Y)=\Hstop$, and $I(X;Z)=\Hroot+\Hbranch$.
This realizes \citeauthor{csiszar1990entropy}'s \citeyearpar{csiszar1990entropy} perfect-graph identity $H(p)=H(G,p)+H(\bar G,p)$ explicitly through \eqref{eq:survivor}.
What our results add to these classical facts is the characterization of every optimal mixture in \cref{thm:bound}, and the streaming reduction of \cref{thm:antichain}, which forces single-history schemes onto $G$ at every step.

\subsection{Corollary 2}
By the chain rule, $I(m;X_{1:T}\mid K)=\sum_{t\le T}I(m;X_t\mid K,X_{<t})\le\sum_{t\le T}H(X_t\mid K,X_{<t})$.
Fix a history $h$.
Then $H(X_t\mid K,X_{<t}=h)=\E_{K\mid h}H(P^K_t(\cdot\mid h))$.
By \cref{thm:antichain}, given $K$ and $h$ each string carries at most one token.
So $H(P^K_t)=H(Q_K)$, where $Q_K$ is the string law of $P^K_t$, and $\supp Q_K$ is a prefix antichain.
By (i), $\E_{K\mid h}Q_K=p_t$.
\Cref{thm:bound} with $\kappa=K$ given $h$ gives $\E_{K\mid h}H(Q_K)\le H(p_t)-\Hstop(p_t)$.
For coin projection, \cref{prop:coin}(d) gives per-step equality when the PRF is modeled as a random function.
\qed

If the receiver recovers $m$ without error, $H(m\mid X_{1:T},K)=0$; since $m$ is independent of $K$, $H(m)=I(m;X_{1:T}\mid K)$ obeys the same bound.
The same application of \cref{thm:bound} gives $I(X_t;K\mid X_{<t}=h)\ge I(\bytes(X_t);K\mid X_{<t}=h)\ge\Hstop(p_t)$, so $I(X_{1:T};K)=\sum_{t\le T}I(X_t;K\mid X_{<t})\ge\sum_{t\le T}\E\Hstop(p_t)$; a finite key of fewer than $\sum_{t\le T}\E\Hstop(p_t)$ bits must therefore be expanded pseudo-randomly (\cref{app:security}).

\subsection{Security against a warden without the key}
\label{app:security}
Each message is sent under the key $K$ with a public \emph{nonce} that never repeats under $K$.
Every pseudo-random value is $\mathrm{PRF}_K(\text{nonce},t,\ell)$, where the label $\ell$ names its use: the stop coins, the homograph draw, or the coder.
As in \cref{def:class}, the sender keeps sampling once the message is embedded.
\begin{proposition}[Security]
\label{prop:security}
Assume that $\mathrm{PRF}$ is a secure pseudo-random function, that the message is uniform and independent of $K$ (in practice a ciphertext), and that the coder is exact: with uniform randomness, its symbol at every step follows the law handed to it, conditionally on all earlier steps.
Let a warden choose prompts and nonces adaptively and receive, for each, the text of the first $T$ tokens of either coin projection or ordinary sampling.
Its advantage in telling the two apart is at most the advantage against $\mathrm{PRF}$ of an adversary that runs the warden and the language model.
\end{proposition}

\noindent\emph{Proof.}
Replace $\mathrm{PRF}_K$ by a uniformly random function.
Distinct labels and fresh nonces give every call a new input, so the coins, the homograph draws and the coder randomness of all steps of all messages become independent and uniform.
Fix a step $t$ and condition on everything before it.
The coins $c$ of step $t$ are independent of this past, and given $c$ the coder emits the string $v$ with probability $\Pt_c(v)$; hence $v$ has probability $\E_c\Pt_c(v)=p_t(v)$ by \cref{prop:coin}(c).
The homograph draw then gives each token $x$ of $v$ its share $\pi_t(x)/p_t(v)$, so $X_t$ follows $\pi_t$ given the past.
Thus the token sequences, and with them the texts, follow ordinary sampling, and the two worlds coincide.
Restoring $\mathrm{PRF}_K$ changes the warden's advantage by at most the advantage of the adversary that simulates the sender with oracle access to $\mathrm{PRF}_K$.
\qed

The proposition is stated for idealized arithmetic and a fixed number of tokens; where a finite stegotext ends is outside its scope.
In our experiments every text has its own key, so the nonce is empty.

\paragraph{Finite precision.}
\label{app:precision}
The range coder works with \texttt{float64} cumulative probabilities and finite-precision rotations, and the coins compare $q_v$ with 53-bit uniforms, so both laws are exact only up to this rounding.
A node stops when its 53-bit uniform times $M(v)$ falls below $p(v)$, which happens with a probability within $2^{-51}$ of the ratio of the \texttt{float64} values of $p(v)$ and $M(v)$.
A path from a root has at most 127 nodes, one per byte length below that of the longest token (128 bytes), so over the at most 136 tokens of a text and with the PRF modeled as a random function, the coins move the law of the tokens by less than $10^{-11}$ in total variation from that under exact coins.
The range coder adds a rounding of its own. Its sequential \texttt{float64} cumulative sums, their normalization and its 53-bit rotation move the law of each token, given the history, by less than $3|\Voc|\cdot2^{-53}<6\cdot10^{-11}$ in total variation from the law handed to it.

\subsection{Filters that ignore the key}
\begin{remark}[Filters that ignore the key]
\label{rem:filter}
Let $w^\ast$ be the largest mass $p(A)$ of a prefix antichain $A$.
Every key-independent law $\bar p$ on a prefix antichain $A$ has $\KL(\bar p\,\|\,p)\ge-\log p(A)\ge-\log w^\ast$, and renormalizing $p$ on a heaviest antichain, as MWIS does, attains both bounds.
Since $w^\ast<1$ whenever two candidates in a prefix relation have positive mass, an exact, zero-error single-history scheme needs a key-dependent antichain, as in coin projection.
\end{remark}

\noindent\emph{Proof.}
Let $p_A=p(\cdot\mid A)$.
For $\bar p$ supported on $A$,
$\KL(\bar p\|p)=\sum_{x\in A}\bar p(x)\log\frac{\bar p(x)}{p(x)}=\KL(\bar p\|p_A)-\log p(A)\ge-\log p(A)$,
with equality if and only if $\bar p=p_A$; maximizing $p(A)$ over antichains gives $-\log w^\ast$.
If two strings $u\pfx v$ both have positive mass, no antichain contains both, so $w^\ast<1$.
For an exact scheme of the class, \cref{thm:antichain} makes every coder input an antichain law, and these laws average to $p$.
If they did not depend on the key, each would equal $p$, whose support is not an antichain.
\qed

\paragraph{Key-conditioned divergence.}
For any mixture with $\bar p=\E_\kappa P_\kappa$,
$\E_\kappa\KL(P_\kappa\|p)=\E_\kappa\sum_x P_\kappa(x)\log\frac{P_\kappa(x)}{\bar p(x)}+\sum_x\bar p(x)\log\frac{\bar p(x)}{p(x)}=I(X;\kappa)+\KL(\bar p\|p)$.
For an exact scheme the second term vanishes, so the coder input under a fixed key differs from $p$ by $\Hstop$ on average under coin projection and by $\Hbranch+\Hstop$ under SyncPool.
A warden without the key never observes the key-conditioned law, so this divergence is not a security loss.

\section{Numerical Checks of the Theory}
\label{app:numerics}

We check the finite-domain identities and bounds of \cref{sec:method} directly in \texttt{float64} arithmetic on 300 fixed-seed random prefix forests with 3 to 10 strings over $\{a,b,c\}^{\le 4}$.
\Cref{tab:numerics} summarizes the residuals:
\begin{itemize}[nosep,leftmargin=*]
\item \textbf{Pointwise subtree conservation (\cref{prop:coin}(b) and \cref{cor:refine}).}
Enumerating every stop pattern $c\in\{0,1\}^{\text{nodes}}$ of every forest confirms that the survivor probability inside the tree of each root $r$ matches $M(r)$ to within $2.22\times 10^{-16}$, and that $H(\Pt_c)\ge\Hroot$ holds for every pattern without exception.
\item \textbf{Stop-entropy upper bound (\cref{thm:bound}).}
Across 6,000 random mixtures $(P_\kappa)$ of antichain-supported string laws with arbitrary mixture weights and mean $\bar p=\E_\kappa P_\kappa$, the largest positive excess $\E_\kappa H(P_\kappa)-(H(\bar p)-\Hstop(\bar p))$ is $1.11\times 10^{-16}$.
\item \textbf{Couplings versus child randomization (\cref{thm:bound} equality condition).}
When incomparable nodes at the same tree depth share a single uniform coin, the marginal equals $p$ to within $10^{-12}$ and $\E_\kappa H(\Pt_\kappa)$ matches $H(p)-\Hstop(p)$ to within $1.33\times 10^{-15}$.
By contrast, a SyncPool-style mixture that encodes the root and uses shared randomness to pick the tree member falls short of $H(p)-\Hstop(p)$ by up to $1.22$ bits whenever $\Hbranch>0$.
\end{itemize}

\paragraph{Along the generated texts.}
\Cref{prop:coin}(d) also holds on the language models of \cref{sec:experiments}.
At every step of every coin-projection text, we record the entropy of the coder input under the coins drawn at that step and the budget $\Hstr-\Hstop$ of the same step.
Summed over the texts of a cell, the first is \res{all}{ident}{ratio}{min}--\res{all}{ident}{ratio}{max} times the second in all \res{all}{ident}{cells}{n} cells.
The range coder turns this entropy into payload bits with an efficiency of \res{all}{ident}{eff}{min}--\res{all}{ident}{eff}{max}, measured as payload bits divided by the summed entropy of the coder input.
The budget of \cref{cor:budget}, not the coder, therefore sets the utilization of coin projection in \cref{tab:main}.

\begin{table}[t]
\centering
\resizebox{\columnwidth}{!}{%
\footnotesize
\setlength{\tabcolsep}{2.5pt}
\begin{tabular}{@{}llc@{}}
\toprule
Property checked & Test domain & Max residual \\
\midrule
$\sum_{v\in T_r}\Pt_c(v)=M(r)$ & All $2^{|\text{nodes}|}$ patterns & $2.22\times 10^{-16}$ \\
$H(\Pt_c)\ge\Hroot$ & All $2^{|\text{nodes}|}$ patterns & $0$ violations \\
$\E_\kappa H(P_\kappa)\le H(\bar p)-\Hstop$ & 6,000 antichain mixtures & $1.11\times 10^{-16}$ \\
Shared-depth coin $\E_\kappa H(\Pt_\kappa)$ & 300 random forests & $1.33\times 10^{-15}$ \\
\bottomrule
\end{tabular}%
}
\caption{Floating-point verification of \cref{prop:coin}, \cref{cor:refine} and \cref{thm:bound} on random prefix forests.}
\label{tab:numerics}
\end{table}

\section{Predicted and Measured Gain}
\label{app:predict}

\Cref{tab:predict} compares, for every model and $k$, the predicted gain of \cref{fig:predict} with two measured ratios of coin projection's to SyncPool's utilization: the ratio of their mean per-text utilizations, and the ratio of their totals, which \cref{tab:main,fig:predict} report.
The total utilization of a cell is the mean of its per-text utilizations weighted by the entropy of each text, so the two ratios differ only through the spread of utilization between texts, which is larger for SyncPool (coefficient of variation up to \res{all}{cv}{syncpool}{max}) than for coin projection (at most \res{all}{cv}{coin}{max}); they differ by at most \res{all}{gain}{excess}{absmax}\%.
The prediction is off by \res{all}{gain}{dev}{total}\% on average and by at most \res{all}{gain}{devmax}{total}\%, in both directions.
It is computed at histories of ordinary sampling (\cref{sec:cost}), not at the histories of the texts that the two methods produced, and it carries the sampling error of those forward passes, which the intervals of \cref{fig:predict} omit.
Computed at the histories of coin projection's own texts (Traj), the same ratio lies inside the 95\% interval of the measured gain in \res{all}{gain}{citraj}{inside} of \res{all}{gain}{cells}{n} cells.

\begin{table*}[t]
\centering
\footnotesize
\setlength{\tabcolsep}{4pt}
\begin{tabular}{@{}l ccccc ccccc ccccc@{}}
\toprule
 & \multicolumn{5}{c}{EN} & \multicolumn{5}{c}{ZH} & \multicolumn{5}{c}{JA} \\
\cmidrule(lr){2-6}\cmidrule(lr){7-11}\cmidrule(lr){12-16}
$k$ & Pred & Traj & Msg & Total & CV & Pred & Traj & Msg & Total & CV & Pred & Traj & Msg & Total & CV \\
\midrule
16 & 1.14 & 1.15 & 1.18 & 1.16 & 0.26 & 1.05 & 1.04 & 1.05 & 1.04 & 0.16 & 1.06 & 1.06 & 1.05 & 1.04 & 0.19 \\
64 & 1.68 & 1.54 & 1.60 & 1.59 & 0.33 & 1.07 & 1.06 & 1.04 & 1.04 & 0.13 & 1.14 & 1.12 & 1.10 & 1.10 & 0.14 \\
256 & 3.16 & 3.21 & 3.38 & 3.37 & 0.41 & 1.08 & 1.07 & 1.09 & 1.09 & 0.11 & 1.24 & 1.18 & 1.17 & 1.17 & 0.16 \\
1024 & 6.43 & 5.86 & 6.34 & 6.20 & 0.65 & 1.11 & 1.10 & 1.10 & 1.09 & 0.12 & 1.31 & 1.29 & 1.29 & 1.31 & 0.22 \\
4096 & 6.20 & 6.08 & 6.93 & 6.80 & 0.60 & 1.16 & 1.17 & 1.15 & 1.16 & 0.16 & 1.60 & 1.60 & 1.58 & 1.59 & 0.24 \\
Full & 6.41 & 6.26 & 6.99 & 6.83 & 0.60 & 2.14 & 2.16 & 2.08 & 2.18 & 0.18 & 2.53 & 2.67 & 2.65 & 2.73 & 0.27 \\
\bottomrule
\end{tabular}

\caption{The predicted gain matches the measured one, most closely when it is computed on the texts actually sent.
Pred is the predicted gain $(\Hstr-\Hstop)/\Hroot$ from the forward passes of \cref{sec:cost}, and Traj the same ratio at the histories of coin projection's texts; Msg and Total are the ratios of coin projection's to SyncPool's mean per-text and total utilization; CV is the coefficient of variation of SyncPool's per-text utilization.}
\label{tab:predict}
\end{table*}

\section{Other Lengths, Sampling Settings and Prompts}
\label{app:robust}

\Cref{tab:robust} repeats Native, SyncPool and coin projection under four changes to the setup of \cref{tab:main}, with the same prompts, keys and streams for all methods: texts of $T=512$ tokens at the full vocabulary (\res{llama}{long}{Full}{n} texts per language), temperature 0.7, nucleus sampling with $p=0.95$, and prompts of 100 words on English and 200 characters on Chinese and Japanese (\res{llama}{temp}{256}{n} texts per cell at $k=256$ and at the full vocabulary).
With a temperature or a nucleus, the modified law is the law of ordinary sampling, and every method starts from it.
SyncPool and coin projection decoded every checked text in all settings.
Coin projection's gain over SyncPool ranged from \res{all}{robust}{gain}{min} to \res{all}{robust}{gain}{max} and, as in \cref{tab:main}, was largest at the full vocabulary, where SyncPool's pool collapses.
Where the prediction of \cref{sec:predict} is available, it lies inside the 95\% interval of the measured gain in \res{all}{robust}{ci}{inside} of \res{all}{robust}{ci}{n} cells.
In the other \res{all}{robust}{ci}{above}, English at the full vocabulary at temperature 0.7 and with long prompts, SyncPool's texts carried only 23 and 48 payload bits on average.
The coder loses a few bits per text whatever the method, a cost that the per-step prediction leaves out; there it took 29\% and 15\% of the entropy that SyncPool's coder saw along its texts, against at most 2\% for coin projection, so the measured gain exceeded the prediction.
At temperature 0.7, SyncPool's texts also had a smaller share of root entropy, 14.4\%, than the fresh samples behind the prediction, 17.3\%; together, the two effects account for the measured gain of 9.00.

Stepwise runs only up to $k=4096$ (\cref{app:details}), so of these settings it ran at $k=256$ with temperature 0.7, with the nucleus and with long prompts, and it decoded every checked text.
Coin projection reached \res{all}{robust}{gainsw}{min}--\res{all}{robust}{gainsw}{max} times Stepwise's utilization, close to its ratio in \cref{tab:main}, and Stepwise again paid with a nonzero divergence.
With a nucleus, the law gives probability zero to every token outside it, while the official Stepwise code widens its candidate set past $k$ when no candidate passes its check and so assumes that every token has positive probability.
When no token of the nucleus passes, we apply the code's own last resort, the most likely token that passes, ranked by the law before truncation, and code no bits on that step.
This happened in \res{all}{robust}{swout}{texts} of the \res{qwen}{nucleus}{256}{n} Chinese texts with $p=0.95$; such a token lies outside the support of ordinary sampling, so that cell's divergence is infinite.

\begin{table*}[t]
\centering
\footnotesize
\setlength{\tabcolsep}{4pt}
\begin{tabular}{@{}lll cccc c cc@{}}
\toprule
 & & & \multicolumn{4}{c}{Util$\uparrow$} & KL$\downarrow$ & \multicolumn{2}{c}{Coin / SyncPool} \\
\cmidrule(lr){4-7}\cmidrule(lr){8-8}\cmidrule(lr){9-10}
Lang & Setting & $k$ & Native$^\ast$ & Stepwise & SyncPool & Coin & Stepwise & Measured & Predicted \\
\midrule
 EN & $T=512$ & Full & .998 & --- & .139 & .904 & --- & 6.49\,{\scriptsize[5.69, 7.37]} & --- \\
 EN & $\tau=0.7$ & 256 & .984 & .992 & .290 & .870 & .0016 & 3.00\,{\scriptsize[2.62, 3.42]} & 3.06 \\
 EN & $\tau=0.7$ & Full & .987 & --- & .102 & .918 & --- & 9.00\,{\scriptsize[6.87, 11.94]} & 5.32 \\
 EN & $p=0.95$ & 256 & .959 & .938 & .469 & .861 & .00047 & 1.84\,{\scriptsize[1.71, 1.99]} & 1.97 \\
 EN & $p=0.95$ & Full & .994 & --- & .412 & .920 & --- & 2.23\,{\scriptsize[2.03, 2.46]} & 2.16 \\
 EN & prompt 100 & 256 & .906 & .904 & .265 & .834 & .00042 & 3.15\,{\scriptsize[2.81, 3.59]} & 3.16 \\
 EN & prompt 100 & Full & .971 & --- & .107 & .897 & --- & 8.36\,{\scriptsize[7.40, 9.46]} & 7.17 \\
\midrule
 ZH & $T=512$ & Full & 1.003 & --- & .397 & .922 & --- & 2.32\,{\scriptsize[2.21, 2.44]} & --- \\
 ZH & $\tau=0.7$ & 256 & .935 & .956 & .823 & .856 & .043 & 1.04\,{\scriptsize[1.00, 1.09]} & 1.06 \\
 ZH & $\tau=0.7$ & Full & .968 & --- & .543 & .902 & --- & 1.66\,{\scriptsize[1.56, 1.77]} & 1.60 \\
 ZH & $p=0.95$ & 256 & .928 & .919 & .800 & .852 & $\infty$ & 1.07\,{\scriptsize[1.04, 1.10]} & 1.08 \\
 ZH & $p=0.95$ & Full & .987 & --- & .803 & .920 & --- & 1.15\,{\scriptsize[1.10, 1.20]} & 1.11 \\
 ZH & prompt 200 & 256 & .854 & .855 & .727 & .780 & .030 & 1.07\,{\scriptsize[1.03, 1.12]} & 1.09 \\
 ZH & prompt 200 & Full & 1.003 & --- & .419 & .912 & --- & 2.18\,{\scriptsize[2.07, 2.28]} & 2.11 \\
\midrule
 JA & $T=512$ & Full & .998 & --- & .321 & .921 & --- & 2.87\,{\scriptsize[2.69, 3.09]} & --- \\
 JA & $\tau=0.7$ & 256 & .923 & .942 & .724 & .882 & .0011 & 1.22\,{\scriptsize[1.13, 1.33]} & 1.23 \\
 JA & $\tau=0.7$ & Full & .948 & --- & .403 & .879 & --- & 2.18\,{\scriptsize[1.95, 2.44]} & 2.11 \\
 JA & $p=0.95$ & 256 & .879 & .914 & .707 & .819 & .0024 & 1.16\,{\scriptsize[1.11, 1.22]} & 1.17 \\
 JA & $p=0.95$ & Full & .985 & --- & .711 & .911 & --- & 1.28\,{\scriptsize[1.21, 1.37]} & 1.31 \\
 JA & prompt 200 & 256 & .827 & .823 & .648 & .754 & .0023 & 1.16\,{\scriptsize[1.12, 1.22]} & 1.15 \\
 JA & prompt 200 & Full & .981 & --- & .337 & .898 & --- & 2.66\,{\scriptsize[2.51, 2.84]} & 2.54 \\
\bottomrule
\end{tabular}

\caption{Coin projection's gain over SyncPool persists with longer texts, other sampling settings and longer prompts.
Util and KL as in \cref{tab:main} (Stepwise run at $k=256$ only); Measured: ratio of the two utilizations with its 95\% paired bootstrap interval; Predicted: $(\Hstr-\Hstop)/\Hroot$ from fresh samples of the same setting (not run for $T=512$); $\infty$: Stepwise emitted a token outside the nucleus.
Prompt: 100 words (EN) or 200 characters (ZH, JA).
$^\ast$Native lost texts in all but three cells, at most 13 in a cell, and \res{llama}{long}{Full}{nativeerr}, \res{qwen}{long}{Full}{nativeerr} and \res{llmjp}{long}{Full}{nativeerr} of \res{llama}{long}{Full}{n} at $T=512$ (\cref{app:rejection}).}
\label{tab:robust}
\end{table*}

\section{Other Coders}
\label{app:coders}

Coin projection changes the law handed to the coder, not the coder, so its gain should not depend on the coder.
\Cref{tab:coders} replaces the range coder by Discop \citep{ding2023discop} with its Huffman recursion and by arithmetic coding \citep{ziegler2019neural}, on the first \res{all}{coder}{any}{n} texts of every cell of \cref{tab:main} with the same prompts, keys and streams.
Our Discop follows the official code rule for rule, except that every tree node draws its uniform from the key and the step, not from a seeded generator, so that the receiver needs no generator state.
Coin projection's gain over SyncPool ranged from \res{all}{coder}{256}{gainmin} to \res{all}{coder}{256}{gainmax} at $k=256$ and from \res{all}{coder}{Full}{gainmin} to \res{all}{coder}{Full}{gainmax} at the full vocabulary over the three coders, and SyncPool and coin projection lost none of the checked texts under any of them.
Discop's Huffman tree carries less than the entropy of each step, so it lowered every method's utilization: coin projection's by \res{all}{coder}{discop@coin}{devabs}\% at most, Native's by \res{all}{coder}{discop@native}{devabs}\% and SyncPool's, whose small pools waste a larger share, by up to \res{all}{coder}{discop@syncpool}{devabs}\%.
Arithmetic coding kept coin projection within \res{all}{coder}{ac@coin}{devabs}\% of its utilization under the range coder.

\begin{table*}[t]
\centering
\footnotesize
\setlength{\tabcolsep}{4pt}
\begin{tabular}{@{}ll ccc c ccc c@{}}
\toprule
 & & \multicolumn{4}{c}{$k=256$} & \multicolumn{4}{c}{Full} \\
\cmidrule(lr){3-6}\cmidrule(lr){7-10}
 & & \multicolumn{3}{c}{Util$\uparrow$} & Coin / & \multicolumn{3}{c}{Util$\uparrow$} & Coin / \\
\cmidrule(lr){3-5}\cmidrule(lr){7-9}
Lang & Coder & Native$^\ast$ & SyncPool & Coin & SyncPool & Native$^\ast$ & SyncPool & Coin & SyncPool \\
\midrule
 \multirow{3}{*}{EN} & Range & .892 & .245 & .812 & 3.31\,{\scriptsize[2.98, 3.65]} & .989 & .131 & .912 & 6.95\,{\scriptsize[6.15, 7.98]} \\
  & Discop & .828 & .215 & .743 & 3.45\,{\scriptsize[3.12, 3.81]} & .917 & .109 & .835 & 7.66\,{\scriptsize[6.70, 8.76]} \\
  & Arithmetic & .870 & .252 & .818 & 3.24\,{\scriptsize[2.88, 3.63]} & .978 & .141 & .906 & 6.42\,{\scriptsize[5.80, 7.10]} \\
\midrule
 \multirow{3}{*}{ZH} & Range & .853 & .734 & .797 & 1.09\,{\scriptsize[1.05, 1.13]} & .989 & .417 & .928 & 2.23\,{\scriptsize[2.12, 2.32]} \\
  & Discop & .791 & .680 & .724 & 1.07\,{\scriptsize[1.04, 1.10]} & .931 & .367 & .856 & 2.33\,{\scriptsize[2.25, 2.42]} \\
  & Arithmetic & .868 & .755 & .791 & 1.05\,{\scriptsize[1.02, 1.08]} & .998 & .422 & .916 & 2.17\,{\scriptsize[2.05, 2.28]} \\
\midrule
 \multirow{3}{*}{JA} & Range & .821 & .648 & .775 & 1.20\,{\scriptsize[1.14, 1.27]} & .980 & .332 & .918 & 2.76\,{\scriptsize[2.54, 3.03]} \\
  & Discop & .762 & .568 & .709 & 1.25\,{\scriptsize[1.18, 1.34]} & .924 & .282 & .850 & 3.01\,{\scriptsize[2.81, 3.23]} \\
  & Arithmetic & .823 & .637 & .771 & 1.21\,{\scriptsize[1.15, 1.27]} & 1.004 & .334 & .929 & 2.79\,{\scriptsize[2.59, 3.00]} \\
\bottomrule
\end{tabular}

\caption{The gain of coin projection over SyncPool holds under each coder.
Util as in \cref{tab:main} on the first \res{all}{coder}{any}{n} texts per cell; Coin / SyncPool: ratio of the two utilizations with its 95\% paired bootstrap interval.
$^\ast$Native's receiver tokenizes $\text{prompt}+\text{text}$ and loses texts under every coder.}
\label{tab:coders}
\end{table*}

\section{Pseudocode}
\label{app:alg}

\Cref{alg:coin} gives one step of coin projection at both ends; \cref{fig:method} illustrates it.

\begin{algorithm}[ht]
\caption{One step of coin projection with a distribution coder. Lines 1--5 are identical at both ends; the public nonce is fresh for every message under $K$ (\cref{app:security}).}
\label{alg:coin}
\small
\begin{algorithmic}[1]
\State $\pi\gets$ LM law on the current history, top-$k$ truncated
\State $p\gets$ string law of $\pi$; masses $M$, stop probabilities $q$
\State $(u_v)_{v\ \mathrm{node}}\gets\mathrm{PRF}_K(\text{nonce},t,\textsf{coin})$ \Comment{one block per node}
\State $c_v\gets\1[u_v<q_v]$ for every node $v$
\State $\Pt\gets$ survivor law \eqref{eq:survivor}
\Statex \textbf{Sender:}
\State $v\gets\textsc{Encode}(\Pt,\text{message state})$
\Statex \textbf{Receiver:}
\State $v\gets$ the survivor that is a prefix of the unread text
\State $\textsc{Decode}(\Pt,v)$
\Statex \textbf{Both:}
\State $x\gets$ token of string $v$, drawn $\propto\pi$ with $\mathrm{PRF}_K(\text{nonce},t,\textsf{hom})$
\end{algorithmic}
\end{algorithm}

\section{Experimental Details}
\label{app:details}

\paragraph{Models, tokenizers and static prefix tables.}
\Cref{tab:models} lists the three base language models and their byte-level vocabulary structures.
For each vocabulary $\Voc$, we extract the raw byte string $\bytes(x)$ of every token $x$: via the byte-level BPE inverse map for Llama-3.1-8B and Qwen3-8B-Base, and via \texttt{<0xHH>} byte-fallback decoding plus replacement of ``\texttt{\spbar}'' (\texttt{U+2581}) by ASCII space for llm-jp-3-7.2b.
Special tokens, including the end of text, added control tokens, and tokens with empty byte strings are masked out before the softmax at every step.
Tokens with identical non-empty byte strings are merged into a single string key for the prefix forest and resolved after string selection via shared randomness (\cref{alg:coin}, line 9).
We precompute all proper byte-prefix pairs $(a,d)$ once per tokenizer and store them sorted both by ascending and by descending ancestor byte length; for an arbitrary top-$k$ subset $S\subseteq\Voc$, finding each candidate's longest prefix in $S$ (its parent in the forest) or shortest prefix in $S$ (its tree root) then requires a single boolean mask over the static pair array and no per-step sorting.
The mask and the scatter-adds that follow touch every pair and every vocabulary entry, so this dense path costs time linear in $|\Voc|$ plus the number of pairs at every $k$.
A sparse path reads only the static ancestors of each string in $S$, through per-string offsets into the pair table, and so runs in $O(|S|)$ time plus the number of pairs whose extension lies in $S$.
It feeds every sum the same terms in the same order and returns bit-identical survivors and laws on all three tokenizers.
Every method that works on the prefix forest (coin projection, SyncPool, hub-drop, Basic and MWIS) takes the sparse path when $16|S|\le|\Voc|$ and the dense path otherwise.

\begin{table}[t]
\centering
\resizebox{\columnwidth}{!}{%
\footnotesize
\setlength{\tabcolsep}{2.5pt}
\begin{tabular}{@{}llccc@{}}
\toprule
Language & Model & Tokenizer & $|\Voc|$ & Prefix pairs \\
\midrule
English & Llama-3.1-8B & Byte-BPE & 128,256 & 458,043 \\
Chinese & Qwen3-8B-Base & Byte-BPE & 151,936 & 519,106 \\
Japanese & llm-jp-3-7.2b & SentencePiece & 99,584 & 245,534 \\
\bottomrule
\end{tabular}%
}
\caption{Models, vocabulary sizes and static proper byte-prefix pair counts.
$|\Voc|$ is the size of the output layer; Llama-3.1-8B, Qwen3-8B-Base and llm-jp-3-7.2b have 128,000, 151,643 and 99,566 regular tokens, and the remaining special or unused entries are masked.}
\label{tab:models}
\end{table}

\paragraph{Corpus and paired design.}
We draw prompts from the English, Chinese and Japanese splits of C4 \citep{raffel2020exploring,xue2021mt5}, taking the first 10 words for English and the first 10 characters for Chinese and Japanese as in \citet{yan2025addressing}.
Prompts containing the Unicode replacement character \texttt{U+FFFD} (11 out of 1,000 in the Chinese shard, 0 in English and Japanese) are discarded deterministically.
For sample index $s$, every method and every truncation size $k\in\{16,64,256,1024,4096,\text{Full}\}$ receives the same prompt, the same uniformly random bit stream, and the same 256-bit secret key $K_s$.
The stop coin of node $v$ at step $t$ is the first 53 bits of the AES-256 block $\mathrm{AES}_{K'}(t\mathbin{\|}v)$, read as a uniform in $[0,1)$, where $K'=\mathrm{HMAC\text{-}SHA256}(K_s,\text{label})$.
Distinct nodes and steps give distinct blocks, so a step computes coins only for its own nodes (at most \res{llama}{tb@cnt}{Full}{nodes} on Llama-3.1-8B; sharing one uniform per tree depth would need at most 85), and their computation is counted in the CPU time Sel of \cref{tab:timing}.
The homograph draw, SyncPool's member draw and the coder take their step uniforms from $\mathrm{HMAC\text{-}SHA256}$ under their own labels.

\paragraph{Uncertainty and complete $k$ grid.}
\Cref{tab:e1-rest} reports the two intermediate truncation sizes ($k=64,1024$) omitted for width in \cref{tab:main}.
Resampling the texts of each cell, each with its own prompt (2,000 bootstrap draws), the 95\% interval of a cell's utilization has a half-width of at most \res{all}{util}{ci}{max}\% of its value (median \res{all}{util}{ci}{median}\%) for every method except SyncPool, whose utilization varies most from text to text (up to \res{all}{util}{ci}{syncpoolmax}\%).
Across all six $k$ settings, the median ratio of the utilization of coin projection to that of Native is \res{all}{ratio}{native}{median}\% (95\% interval \res{all}{ratio}{native}{medlo}--\res{all}{ratio}{native}{medhi}\%, paired by text), and to that of Stepwise \res{all}{ratio}{stepwise}{median}\% (\res{all}{ratio}{stepwise}{medlo}--\res{all}{ratio}{stepwise}{medhi}\%).
BPT also depends on the entropy of each method's own text, so one method can have the higher BPT and the lower utilization in the same cell; Basic and Native on English at $k=16$ (BPT \res{llama}{basic}{16}{bpt} vs.\ \res{llama}{native}{16}{bpt}, Util \res{llama}{basic}{16}{util} vs.\ \res{llama}{native}{16}{util} in \cref{tab:main}) are an example.
Utilization divides by the entropy that each text actually offered and is the measure we compare.
A method that distorts the law also moves its text to other histories. On English at $k=4096$, Basic's texts offer about 1.5 times the entropy per token of the other methods' (BPT over Util in \cref{tab:main}).
The payload is a realized quantity and the entropy an expectation, so on a finite sample utilization can exceed 1, as it does for Native in two cells of \cref{tab:robust}.

\begin{table}[t]
\centering
\resizebox{\columnwidth}{!}{%
\footnotesize
\setlength{\tabcolsep}{2.2pt}
\begin{tabular}{@{}ll ccc ccc@{}}
\toprule
 &  & \multicolumn{3}{c}{$k=64$} & \multicolumn{3}{c}{$k=1024$} \\
\cmidrule(lr){3-5}\cmidrule(lr){6-8}
Model & Method & BPT$\uparrow$ & Util$\uparrow$ & $\KL_{\mathrm{sec}}\downarrow$ & BPT$\uparrow$ & Util$\uparrow$ & $\KL_{\mathrm{sec}}\downarrow$ \\
\midrule
 \multirow{8}{*}{\shortstack[l]{Llama-3.1-8B\\(EN, 128k)}} & Native$^\ast$ & 2.65 & .794 & 0 & 3.51 & .944 & 0 \\
  & Basic & 3.25 & .815 & .718 & 5.22 & .984 & 1.05 \\
  & MWIS & 2.45 & .735 & .064 & 3.14 & .878 & .083 \\
  & Stepwise & 2.73 & .798 & .0012 & 3.47 & .945 & .0043 \\
  & Hub-drop & 2.37 & .697 & .0032 & 2.89 & .799 & .0058 \\
\cmidrule(lr){2-8}
  & SyncPool & 1.57 & .459 & 0 & 0.52 & .139 & 0 \\
  & LAS & 2.68 & .785 & 0 & 3.44 & .923 & 0 \\
  & \textbf{Coin (ours)} & 2.52 & \textbf{.729} & 0 & 3.21 & \textbf{.861} & 0 \\
\midrule
 \multirow{8}{*}{\shortstack[l]{Qwen3-8B-Base\\(ZH, 152k)}} & Native$^\ast$ & 2.28 & .764 & 0 & 3.32 & .937 & 0 \\
  & Basic & 3.12 & .805 & .739 & 4.78 & .930 & .792 \\
  & MWIS & 2.09 & .696 & .059 & 2.86 & .846 & .082 \\
  & Stepwise & 2.41 & .750 & .021 & 3.43 & .934 & .024 \\
  & Hub-drop & 1.98 & .675 & .0016 & 2.74 & .811 & .0024 \\
\cmidrule(lr){2-8}
  & SyncPool & 2.05 & .664 & 0 & 2.86 & .788 & 0 \\
  & LAS & 2.31 & .758 & 0 & 3.34 & .932 & 0 \\
  & \textbf{Coin (ours)} & 2.18 & \textbf{.690} & 0 & 3.09 & \textbf{.862} & 0 \\
\midrule
 \multirow{8}{*}{\shortstack[l]{llm-jp-3-7.2b\\(JA, 100k)}} & Native$^\ast$ & 2.36 & .737 & 0 & 3.47 & .911 & 0 \\
  & Basic & 2.77 & .733 & .596 & 4.12 & .904 & .698 \\
  & MWIS & 2.10 & .676 & .059 & 2.98 & .821 & .084 \\
  & Stepwise & 2.33 & .736 & .0022 & 3.46 & .911 & .0021 \\
  & Hub-drop & 1.97 & .644 & .0018 & 2.79 & .776 & .0032 \\
\cmidrule(lr){2-8}
  & SyncPool & 1.93 & .604 & 0 & 2.42 & .640 & 0 \\
  & LAS & 2.31 & .716 & 0 & 3.34 & .912 & 0 \\
  & \textbf{Coin (ours)} & 2.06 & \textbf{.664} & 0 & 3.12 & \textbf{.836} & 0 \\
\bottomrule
\end{tabular}
}
\caption{Results at the intermediate truncation sizes $k=64$ and $k=1024$ complementing \cref{tab:main} ($n=\res{all}{e1}{any}{nmin}$ paired texts per cell; $^\ast$Native lost \res{llama}{native}{64}{err}, \res{qwen}{native}{64}{err} and \res{llmjp}{native}{64}{err} at $k=64$, and \res{llama}{native}{1k}{err}, \res{qwen}{native}{1k}{err} and \res{llmjp}{native}{1k}{err} at $k=1024$ on EN, ZH and JA; \cref{tab:native-err}).}
\label{tab:e1-rest}
\end{table}

\paragraph{Byte-level channel and UTF-8 completion.}
The channel carries only the continuation as a Unicode string.
With byte-level tokenizers, the 128th token can end inside a multi-byte UTF-8 character.
The sender then samples up to 8 more tokens, embedding as before, until the bytes are valid UTF-8; the payload is the stream prefix fixed by all tokens sent.
Bytes that can no longer become valid, such as a lead byte followed by a token that does not continue it, cannot be sent as Unicode at all.
Ordinary sampling produces them with the same probability, and the sender sees them and handles the text as a user of ordinary sampling would, for example by sampling a new one, so the event reveals nothing.
Such a text counts as carrying no payload.
Whether a text is extended depends only on its bytes, so a stegotext of an exact method follows ordinary sampling under the same length rule as its cover, and its length reveals nothing beyond the text.
Every text of an exact method was valid UTF-8 within the 8 extra tokens; four texts of filters (Basic on Chinese, MWIS and hub-drop on Japanese) did not, and they count as carrying no payload.

\paragraph{Baseline implementations and equivalence checks.}
To separate algorithmic behavior from string-decoding artifacts, the prefix-based methods operate on the exact byte-level prefix graph:
\begin{itemize}[nosep,leftmargin=*]
\item \emph{Basic} \citep{nozaki2022addressing}, \emph{MWIS} \citep{yan2023secure}, and \emph{SyncPool} \citep{qi2024provably} call the official pool-construction functions of \citet{yan2025addressing} with token strings replaced by exact byte strings $\bytes(x)$.
On 24 real histories of each of the three language models and $k\in\{16,64,256\}$, our byte-level builders return the official repository's candidate sets with identical weights for Basic and MWIS and group masses equal to within a relative $10^{-12}$ for SyncPool.
\item \emph{Exact MWIS.} The official code finds the maximum-weight independent set greedily, as \citet{yan2023secure} do for general graphs.
On a prefix conflict graph, an independent set is an antichain of the prefix forest, and the heaviest one follows from the tree recursion $\mathrm{best}(v)=\max\bigl(p(v),\sum_{c\in\ch(v)}\mathrm{best}(c)\bigr)$, which we evaluate exactly at every $k$, including the full vocabulary.
On the same 24 histories at $k=256$, the greedy set is lighter than the exact one on 12 to 20 histories per model and adds up to 0.24 nats of divergence, so exact MWIS is the stronger baseline; by \cref{rem:filter}, no key-independent filter that keeps a prefix-free set has a smaller divergence.
\item \emph{Stepwise verification} \citep{yan2025addressing} calls the official candidate check unchanged.
The check keeps a string $v$ together with its extensions, so the message makes the stop decisions, and after $v$ it rejects every token after which the history is no longer canonically tokenized, such as one that continues $v$ into a longer token.
In the example of the remark in \cref{app:proofs}, this is the case $\lambda=1$, which is exact only if $\eta=0$; each step lies $-\log(1-r)$ nats from top-$k$ sampling, where $r$ is the rejected mass.
Since the check compares text, it also rejects tokens that end inside a UTF-8 character, which make up \res{qwen}{swdiag}{any}{sharemin}--\res{qwen}{swdiag}{any}{sharemax}\% of its rejections on Chinese and thus most of the mass it removes there.
\citet{yan2025addressing} identify these partial-character tokens as the main source of the temporary inconsistencies they observe on Chinese and Japanese, and their check removes them by design.
\item \emph{Range of $k$.} Basic and Stepwise run up to $k=4096$, where the grid of the public evaluation script of \citet{yan2025addressing} ends; the other methods of \cref{tab:main} also run at the full vocabulary.
Stepwise's check retokenizes the history once for every candidate, so its time grows in proportion to $k$: on English, from \res{llama}{time@stepwise}{256}{sel}\,ms per token at $k=256$ to \res{llama}{time@stepwise}{4k}{sel}\,ms at $k=4096$ (\cref{tab:timing}).
The full vocabularies hold 24 to 37 times as many regular tokens as $k=4096$ (\cref{tab:models}), so at this rate Stepwise would take 10 to 18\,s per token, or 1.6 to 2.6 days of sender time for the \res{all}{e1}{any}{nmin} texts of one cell.
Basic, which shows how a filter trades divergence for capacity, already lies \res{llmjp}{basic}{4k}{kl}--\res{llama}{basic}{4k}{kl} nats per step from top-$k$ sampling at $k=4096$ (\cref{tab:main}).
\item \emph{Range coding.} All methods, LAS included, share our implementation of the rotation range coder of \citet{yan2026efficient} (their Algorithms 3 and 4), with a step uniform per text and step.
Its rotation makes the symbol at every step follow the law it is handed, so it is an exact coder in the sense of \cref{prop:security}.
The receiver recovers the payload as the common prefix of all stream values consistent with the decoded symbols.
\item \emph{Method-matched receivers.}
The public evaluation script of \citet{yan2025addressing} applies canonical retokenization to all baselines and draws SyncPool members with an unseeded generator, which artificially inflates extraction failures for Basic, MWIS and SyncPool.
In our evaluation, Native and Stepwise use canonical retokenization of $\text{prompt}+\text{text}$, and when this changes the prompt tokens, Native falls back to tokenizing the text alone; Basic and MWIS parse the unique matching byte prefix in the prefix-free candidate set; SyncPool and hub-drop parse the unique pool root and draw the pool member with the shared step uniform; coin projection parses the unique surviving prefix in $\supp\Pt_c$; and LAS uses the receiver of its public release, which keeps the token histories consistent with the text.
\end{itemize}

\paragraph{Hardware.}
All runs use one NVIDIA GeForce RTX 5090 (32\,GB) with \texttt{bfloat16} model weights, and \texttt{float64} CPU code for the coder, the projection and the baselines on 8 threads of an Intel Xeon Platinum 8470Q.
\Cref{app:timing} describes how we timed them.

\paragraph{Parallel receivers.}
Non-canonical tokenizations are rare (\cref{sec:cost}).
A receiver could therefore guess the canonical tokenization, verify it with one block forward pass, and fall back at the first deviation.
This works only if sender and receiver run with an identical, fixed computation shape \citep{yan2026invocation,qi2026blocklock,zhao2026corun,qin2026vosti}.
The idea applies to every single-history scheme, so we do not use it in our experiments; all methods there decode sequentially.

\section{Timing}
\label{app:timing}

\Cref{tab:timing} reports the time per token of every method, LAS included, on the first 10 texts of each model and $k$, with diagnostics switched off.
The methods take turns text by text, so a slow drift of the machine affects them alike, and one untimed text per method and $k$ warms up the model and the caches.
Times are totals over the texts divided by their tokens.
\emph{Fwd} is the model's forward pass, synchronized with the GPU before the clock is read, and counts all of LAS's forward passes; \emph{Sel} is the rest of the sender's work per token on the CPU: softmax and top-$k$ selection, the method's plan (the step that turns the top-$k$ law into the coder input) and the coder step; \emph{Recv} is the receiver's whole time per token, forward passes included.
The ratios of LAS's time to coin projection's in the main text use the sender's time, Fwd plus Sel.
Dividing the bits per token of \cref{tab:main} by this time, and taking each method at its highest-throughput $k$ among the four in \cref{tab:timing}, coin projection sent 174, 133 and 191 payload bits per second on English, Chinese and Japanese, against 93, 110 and 134 for SyncPool, 68, 68 and 76 for LAS as released, without batching, and 200, 151 and 215 for Native.
SyncPool's throughput is highest at $k=16$, 4096 and 256, where its CPU work takes 2.3--5.7\,ms per token; if it were charged only the CPU time of ordinary sampling there, coin projection would still send 1.11--1.85 times its bits per second, so the gain in throughput comes from capacity rather than from SyncPool's implementation.
Coin projection's is highest at $k=4096$, where it keeps 88--95\% of its bits per token at the full vocabulary.
Apart from Native's, every receiver decoded these texts correctly.

\Cref{tab:tb} isolates the work that differs between methods.
On 10 of the histories of \cref{tab:gof} per model, it times the plan and the coder step 20 times at every $k$, each time with a fresh key and step, with the methods in rotated order, and reports the median over the repetitions and then over the histories.
The sizes this work depends on are the number $|S|$ of distinct strings in the candidate set, the prefix pairs among them, and the nodes, the strings of $S$ that are proper prefixes of other strings of $S$ and thus need a stop coin.
With the sparse path (\cref{app:details}), coin projection's step at $k=16$ takes \res{llama}{tb@coin}{16}{sparse}\,ms on English against \res{llama}{tb@coin}{16}{dense}\,ms on the dense path; \cref{tab:tb} reports the path that the rule selects, timed in repetitions of its own.
At the full vocabulary, it takes \res{llama}{tb@coin}{Full}{auto}, \res{qwen}{tb@coin}{Full}{auto} and \res{llmjp}{tb@coin}{Full}{auto}\,ms on English, Chinese and Japanese, against \res{llama}{tb@syncpool}{Full}{auto}, \res{qwen}{tb@syncpool}{Full}{auto} and \res{llmjp}{tb@syncpool}{Full}{auto}\,ms for SyncPool's official pool construction.
By \cref{cor:refine}, SyncPool's coder input also follows from the forest pass that coin projection uses, so this gap reflects the implementations rather than the designs.
Exact MWIS on the same forest takes \res{llama}{tb@mwis}{Full}{auto}, \res{qwen}{tb@mwis}{Full}{auto} and \res{llmjp}{tb@mwis}{Full}{auto}\,ms at the full vocabulary.
Stepwise, whose check retokenizes the history once per candidate (\cref{app:details}), takes \res{llama}{tb@stepwise}{4k}{auto}, \res{qwen}{tb@stepwise}{4k}{auto} and \res{llmjp}{tb@stepwise}{4k}{auto}\,ms already at $k=4096$, against \res{llama}{tb@coin}{4k}{auto}, \res{qwen}{tb@coin}{4k}{auto} and \res{llmjp}{tb@coin}{4k}{auto}\,ms for coin projection.

\begin{table*}[t]
\centering
\footnotesize
\setlength{\tabcolsep}{3.2pt}
\begin{tabular}{@{}ll rrr rrr rrr rrr@{}}
\toprule
 & & \multicolumn{3}{c}{$k=16$} & \multicolumn{3}{c}{$k=256$} & \multicolumn{3}{c}{$k=4096$} & \multicolumn{3}{c}{Full} \\
\cmidrule(lr){3-5}\cmidrule(lr){6-8}\cmidrule(lr){9-11}\cmidrule(lr){12-14}
Model & Method & Fwd & Sel & Recv & Fwd & Sel & Recv & Fwd & Sel & Recv & Fwd & Sel & Recv \\
\midrule
 \multirow{8}{*}{\shortstack[l]{Llama-3.1-8B\\(EN)}} & Native$^\ast$ & 14.9 & 2.2 & 17.0 & 14.8 & 2.3 & 17.1 & 14.8 & 3.2 & 18.1 & 14.9 & 9.3 & 24.5 \\
  & Basic & 14.9 & 2.4 & 17.3 & 14.8 & 3.0 & 17.9 & 15.0 & 27.2 & 42.3 & --- & --- & --- \\
  & MWIS & 14.9 & 2.4 & 17.4 & 14.9 & 2.7 & 17.6 & 14.9 & 5.5 & 20.4 & 15.2 & 74.3 & 89.2 \\
  & Stepwise & 14.9 & 4.2 & 19.1 & 14.9 & 32.7 & 47.6 & 15.3 & 515.1 & 528.9 & --- & --- & --- \\
  & Hub-drop & 14.9 & 2.5 & 17.4 & 14.9 & 2.8 & 17.7 & 14.9 & 6.7 & 21.6 & 15.2 & 122.4 & 137.9 \\
\cmidrule(lr){2-14}
  & SyncPool & 14.9 & 2.4 & 17.4 & 14.9 & 2.6 & 17.6 & 15.0 & 5.1 & 20.1 & 15.0 & 73.5 & 88.4 \\
  & LAS & 23.1 & 6.1 & 30.2 & 44.5 & 13.8 & 59.8 & 67.8 & 34.3 & 104.5 & 72.2 & 66.3 & 143.1 \\
  & \textbf{Coin (ours)} & 14.9 & 2.4 & 17.3 & 14.9 & 2.6 & 17.5 & 15.0 & 4.2 & 19.2 & 15.6 & 22.5 & 38.1 \\
\midrule
 \multirow{8}{*}{\shortstack[l]{Qwen3-8B-Base\\(ZH)}} & Native$^\ast$ & 21.2 & 2.5 & 22.7 & 21.2 & 2.6 & 23.9 & 21.3 & 3.5 & 24.9 & 21.2 & 9.5 & 31.0 \\
  & Basic & 21.1 & 2.7 & 23.9 & 21.3 & 3.2 & 24.7 & 21.4 & 23.1 & 44.3 & --- & --- & --- \\
  & MWIS & 21.2 & 2.7 & 24.0 & 21.2 & 3.0 & 24.3 & 21.2 & 5.3 & 26.5 & 21.6 & 83.9 & 105.3 \\
  & Stepwise & 21.1 & 4.4 & 25.5 & 21.3 & 33.0 & 54.3 & 21.6 & 478.6 & 500.7 & --- & --- & --- \\
  & Hub-drop & 21.2 & 2.8 & 24.2 & 21.2 & 3.1 & 24.4 & 21.2 & 6.5 & 27.8 & 21.8 & 142.8 & 164.8 \\
\cmidrule(lr){2-14}
  & SyncPool & 21.1 & 2.7 & 24.0 & 21.3 & 3.0 & 24.3 & 21.3 & 5.7 & 27.2 & 21.4 & 87.0 & 108.4 \\
  & LAS & 26.6 & 5.7 & 33.4 & 32.8 & 9.3 & 43.3 & 49.3 & 45.3 & 97.2 & 88.6 & 64.3 & 155.3 \\
  & \textbf{Coin (ours)} & 21.2 & 2.7 & 24.1 & 21.2 & 2.9 & 24.2 & 21.3 & 4.3 & 25.6 & 21.5 & 24.1 & 45.7 \\
\midrule
 \multirow{8}{*}{\shortstack[l]{llm-jp-3-7.2b\\(JA)}} & Native$^\ast$ & 14.5 & 1.9 & 15.5 & 14.6 & 2.0 & 16.5 & 14.5 & 3.0 & 17.5 & 14.6 & 7.6 & 22.4 \\
  & Basic & 14.6 & 2.1 & 16.6 & 14.6 & 2.8 & 17.4 & 14.7 & 23.8 & 38.4 & --- & --- & --- \\
  & MWIS & 14.6 & 2.1 & 16.6 & 14.5 & 2.3 & 16.9 & 14.7 & 4.8 & 19.3 & 14.7 & 45.3 & 59.7 \\
  & Stepwise & 14.5 & 3.5 & 18.0 & 14.6 & 27.0 & 41.3 & 14.9 & 433.5 & 447.6 & --- & --- & --- \\
  & Hub-drop & 14.5 & 2.2 & 16.8 & 14.6 & 2.6 & 17.2 & 14.6 & 6.6 & 21.1 & 14.7 & 88.6 & 103.3 \\
\cmidrule(lr){2-14}
  & SyncPool & 14.5 & 2.1 & 16.7 & 14.6 & 2.3 & 16.9 & 14.6 & 4.9 & 19.5 & 14.6 & 52.6 & 67.1 \\
  & LAS & 19.4 & 4.3 & 24.5 & 29.1 & 8.9 & 39.1 & 54.0 & 55.4 & 112.4 & 65.1 & 52.1 & 120.2 \\
  & \textbf{Coin (ours)} & 14.6 & 2.1 & 16.6 & 14.6 & 2.2 & 16.9 & 14.6 & 3.8 & 18.4 & 14.9 & 15.3 & 30.2 \\
\bottomrule
\end{tabular}

\caption{Coin projection adds little to the forward pass, and LAS pays for its extra forward passes.
Milliseconds per token on 10 texts per cell with one RTX 5090; Fwd: model forward passes; Sel: softmax, top-$k$, plan and coder step on the CPU; Recv: the receiver, forward passes included.
Basic and Stepwise were not run at the full vocabulary (\cref{app:details}).
$^\ast$Native loses texts to retokenization (\cref{tab:main}).}
\label{tab:timing}
\end{table*}

\begin{table*}[t]
\centering
\footnotesize
\setlength{\tabcolsep}{3.2pt}
\begin{tabular}{@{}ll rrr r r r r r r r@{}}
\toprule
 & & \multicolumn{3}{c}{Sizes (median)} & \multicolumn{7}{c}{Plan and coder step (ms, median)} \\
\cmidrule(lr){3-5}\cmidrule(lr){6-12}
 & $k$ & $|S|$ & Pairs & Nodes & Native$^\ast$ & Basic & MWIS & Stepwise & Hub-drop & SyncPool & \textbf{Coin (ours)} \\
\midrule
 \multirow{6}{*}{EN} & 16 & 16 & 1 & 1 & 0.78 & 0.94 & 1.01 & 1.65 & 1.07 & 1.01 & 1.07 \\
  & 64 & 64 & 45 & 7 & 0.81 & 1.04 & 1.07 & 3.63 & 1.14 & 1.06 & 1.10 \\
  & 256 & 256 & 355 & 44 & 0.89 & 1.51 & 1.27 & 11.39 & 1.38 & 1.21 & 1.24 \\
  & 1024 & 1{,}024 & 2{,}294 & 234 & 1.16 & 3.76 & 1.88 & 43.73 & 2.17 & 1.77 & 1.66 \\
  & 4096 & 4{,}096 & 12{,}413 & 1{,}139 & 1.98 & --- & 4.14 & 172.93 & 5.40 & 3.78 & 3.04 \\
  & Full & 128{,}000 & 458{,}043 & 41{,}302 & 7.95 & --- & 73.97 & --- & 122.50 & 75.88 & 22.06 \\
\midrule
 \multirow{6}{*}{ZH} & 16 & 16 & 1 & 1 & 0.85 & 1.00 & 1.09 & 1.61 & 1.19 & 1.08 & 1.14 \\
  & 64 & 64 & 7 & 6 & 0.86 & 1.09 & 1.13 & 3.21 & 1.22 & 1.12 & 1.18 \\
  & 256 & 256 & 60 & 36 & 0.95 & 1.46 & 1.32 & 9.67 & 1.44 & 1.31 & 1.30 \\
  & 1024 & 1{,}024 & 417 & 190 & 1.30 & 3.07 & 1.90 & 35.39 & 2.24 & 2.03 & 1.73 \\
  & 4096 & 4{,}096 & 3{,}340 & 890 & 1.88 & --- & 3.46 & 143.61 & 4.77 & 3.97 & 2.74 \\
  & Full & 151{,}643 & 519{,}106 & 45{,}399 & 7.86 & --- & 83.63 & --- & 129.86 & 88.83 & 23.47 \\
\midrule
 \multirow{6}{*}{JA} & 16 & 16 & 5 & 2 & 0.71 & 0.87 & 0.92 & 1.47 & 1.02 & 0.93 & 0.99 \\
  & 64 & 64 & 38 & 10 & 0.73 & 0.98 & 0.98 & 3.08 & 1.10 & 0.99 & 1.02 \\
  & 256 & 256 & 198 & 39 & 0.83 & 1.40 & 1.17 & 9.73 & 1.41 & 1.18 & 1.16 \\
  & 1024 & 1{,}024 & 1{,}077 & 185 & 1.22 & 3.26 & 1.82 & 36.55 & 2.39 & 1.88 & 1.64 \\
  & 4096 & 4{,}096 & 6{,}006 & 871 & 1.85 & --- & 3.51 & 148.21 & 5.44 & 3.84 & 2.71 \\
  & Full & 99{,}441 & 245{,}534 & 23{,}850 & 6.62 & --- & 43.70 & --- & 96.58 & 59.36 & 14.54 \\
\bottomrule
\end{tabular}

\caption{At the full vocabulary, coin projection has the fastest plan and coder step among the methods on the prefix forest.
Medians over 10 fixed histories per model and 20 repetitions; $|S|$: distinct strings in the candidate set; Pairs: proper prefix pairs within $S$; Nodes: strings of $S$ with an extension in $S$, one stop coin each.
$^\ast$Native loses texts to retokenization (\cref{tab:main}).}
\label{tab:tb}
\end{table*}

\section{Best Truncation and Goodness of Fit}
\label{app:bestk-gof}

\Cref{tab:bestk} compares each method at the truncation size $k$ where its BPT is highest.
With both at their best $k$, coin projection's BPT is \res{all}{bestk}{syncpool}{min}--\res{all}{bestk}{syncpool}{max} times SyncPool's, and it also exceeds that of exact MWIS and hub-drop.
Native and LAS reach a slightly higher BPT, Native at the cost of its failures and LAS at the cost of its extra forward passes; Basic reaches a higher BPT and Stepwise a similar one, both with a divergence from ordinary sampling.
Both run only up to $k=4096$ (\cref{app:details}), and Stepwise peaks there, as Basic does on English and Japanese.

\Cref{tab:gof} tests the next-token law directly.
Each model has 20 fixed histories, each a prompt followed by 0, 8, 16 or 24 tokens of ordinary sampling.
At each history and $k$, we draw 5,000 next tokens through each pipeline (plan, coder step with range coding, homograph draw), every draw with its own key and its own uniform 128-bit message, and compare their counts with the top-$k$ law by a Pearson $\chi^2$ test, pooling the tokens with expected counts below 5.
Fisher's method combines the 20 $p$-values of a cell.
An exact method should be rejected at the 5\% level on about one history in 20; Native, SyncPool and coin projection were rejected in \res{all}{gof}{exact}{rejtot} of \res{all}{gof}{exact}{ctxtot} tests, and the smallest of their 27 combined $p$-values, \res{all}{gof}{exact}{fishermin}, is what the smallest of 27 uniform draws is on average ($1/28$).
Basic and MWIS change the law and act as positive controls.
Stepwise is not tested, since its per-step divergence follows from the mass its check rejects (\cref{app:details}) and \cref{tab:main} reports it.

\begin{table}[t]
\centering
\resizebox{\columnwidth}{!}{%
\footnotesize
\setlength{\tabcolsep}{2.4pt}
\begin{tabular}{@{}ll c c c c@{}}
\toprule
Model & Method & best $k$ & BPT$\uparrow$ & $\KL_{\mathrm{sec}}\downarrow$ & Coin / method (BPT) \\
\midrule
 \multirow{8}{*}{\shortstack[l]{Llama-3.1-8B\\(EN)}} & Native$^\ast$ & Full & 3.79 & 0 & 0.93\,{\scriptsize[0.88, 0.98]} \\
  & Basic & 4k & 5.55 & 1.02 & 0.63\,{\scriptsize[0.58, 0.69]} \\
  & MWIS & 4k & 3.34 & .088 & 1.05\,{\scriptsize[1.00, 1.10]} \\
  & Stepwise & 4k & 3.69 & .0043 & 0.95\,{\scriptsize[0.91, 1.00]} \\
  & Hub-drop & Full & 3.20 & .0073 & 1.10\,{\scriptsize[1.04, 1.16]} \\
\cmidrule(lr){2-6}
  & SyncPool & 16 & 1.61 & 0 & 2.18\,{\scriptsize[2.04, 2.35]} \\
  & LAS & Full & 3.79 & 0 & 0.93\,{\scriptsize[0.89, 0.97]} \\
  & \textbf{Coin (ours)} & Full & 3.52 & 0 & 1 \\
\midrule
 \multirow{8}{*}{\shortstack[l]{Qwen3-8B-Base\\(ZH)}} & Native$^\ast$ & Full & 4.05 & 0 & 0.95\,{\scriptsize[0.86, 1.04]} \\
  & Basic & 1k & 4.78 & .792 & 0.80\,{\scriptsize[0.73, 0.89]} \\
  & MWIS & Full & 3.23 & .091 & 1.19\,{\scriptsize[1.10, 1.29]} \\
  & Stepwise & 4k & 3.97 & .022 & 0.97\,{\scriptsize[0.90, 1.05]} \\
  & Hub-drop & 4k & 3.26 & .0028 & 1.18\,{\scriptsize[1.08, 1.29]} \\
\cmidrule(lr){2-6}
  & SyncPool & 4k & 2.99 & 0 & 1.29\,{\scriptsize[1.16, 1.43]} \\
  & LAS & Full & 4.02 & 0 & 0.96\,{\scriptsize[0.87, 1.04]} \\
  & \textbf{Coin (ours)} & Full & 3.85 & 0 & 1 \\
\midrule
 \multirow{8}{*}{\shortstack[l]{llm-jp-3-7.2b\\(JA)}} & Native$^\ast$ & Full & 4.03 & 0 & 0.97\,{\scriptsize[0.90, 1.04]} \\
  & Basic & 4k & 4.39 & .708 & 0.89\,{\scriptsize[0.82, 0.96]} \\
  & MWIS & Full & 3.37 & .097 & 1.16\,{\scriptsize[1.06, 1.26]} \\
  & Stepwise & 4k & 3.65 & .003 & 1.07\,{\scriptsize[0.99, 1.15]} \\
  & Hub-drop & Full & 3.55 & .0046 & 1.10\,{\scriptsize[1.02, 1.19]} \\
\cmidrule(lr){2-6}
  & SyncPool & 1k & 2.42 & 0 & 1.61\,{\scriptsize[1.48, 1.74]} \\
  & LAS & Full & 4.01 & 0 & 0.97\,{\scriptsize[0.90, 1.06]} \\
  & \textbf{Coin (ours)} & Full & 3.90 & 0 & 1 \\
\bottomrule
\end{tabular}
}
\caption{At each method's best $k$, coin projection exceeds SyncPool in BPT on all three models.
Best $k$: the $k$ with the highest BPT, among $k\le4096$ for Basic and Stepwise; last column: ratio of coin projection's BPT at its best $k$ to the method's at its best $k$, with its 95\% paired bootstrap interval.
$^\ast$Native loses texts to retokenization (\cref{tab:main}).}
\label{tab:bestk}
\end{table}

\begin{table}[t]
\centering
\resizebox{\columnwidth}{!}{%
\footnotesize
\setlength{\tabcolsep}{2.6pt}
\begin{tabular}{@{}ll rr rr rr@{}}
\toprule
 & & \multicolumn{2}{c}{$k=16$} & \multicolumn{2}{c}{$k=256$} & \multicolumn{2}{c}{Full} \\
\cmidrule(lr){3-4}\cmidrule(lr){5-6}\cmidrule(lr){7-8}
Lang & Method & Rej. & Fisher $p$ & Rej. & Fisher $p$ & Rej. & Fisher $p$ \\
\midrule
 \multirow{5}{*}{EN} & Native$^\ast$ & 3/20 & 0.24 & 0/20 & 0.57 & 1/20 & 0.6 \\
  & Basic & 14/20 & 0 & 18/20 & 0 & --- & --- \\
  & MWIS & 13/20 & 0 & 16/20 & 0 & 15/20 & 0 \\
  & SyncPool & 1/20 & 0.45 & 1/20 & 0.78 & 0/20 & 1 \\
  & \textbf{Coin (ours)} & 0/20 & 0.96 & 1/20 & 0.47 & 0/20 & 0.55 \\
\midrule
 \multirow{5}{*}{ZH} & Native$^\ast$ & 1/20 & 0.14 & 2/20 & 0.28 & 1/20 & 0.7 \\
  & Basic & 17/20 & 0 & 19/20 & 0 & --- & --- \\
  & MWIS & 15/20 & 0 & 19/20 & 0 & 20/20 & 0 \\
  & SyncPool & 2/20 & 0.14 & 1/20 & 0.43 & 0/20 & 0.84 \\
  & \textbf{Coin (ours)} & 0/20 & 0.49 & 0/20 & 0.89 & 0/20 & 0.79 \\
\midrule
 \multirow{5}{*}{JA} & Native$^\ast$ & 3/20 & 0.11 & 2/20 & 0.056 & 2/20 & 0.34 \\
  & Basic & 15/20 & 0 & 18/20 & 0 & --- & --- \\
  & MWIS & 15/20 & 0 & 16/20 & 0 & 16/20 & 0 \\
  & SyncPool & 1/20 & 0.82 & 1/20 & 0.55 & 3/20 & 0.11 \\
  & \textbf{Coin (ours)} & 0/20 & 0.97 & 1/20 & 0.27 & 1/20 & 0.036 \\
\bottomrule
\end{tabular}
}
\caption{Exact methods pass a goodness-of-fit test on real histories, and the filters fail it.
Rej.: histories (of 20) rejected at the 5\% level; Fisher $p$: combined $p$-value of the 20 tests.
Basic was not run at the full vocabulary (\cref{app:details}).
$^\ast$Native's emitted law is exact; its failures occur at the receiver.}
\label{tab:gof}
\end{table}

\section{Native Failures and Ambiguity Rates}
\label{app:native-err}

Across the \res{all}{recvall}{coin}{decoded} Native texts in \cref{tab:main,tab:e1-rest} (\res{all}{native}{any}{n} per language), \res{all}{native}{all}{err} decode wrongly (\res{llama}{native}{all}{err} on English, \res{qwen}{native}{all}{err} on Chinese and \res{llmjp}{native}{all}{err} on Japanese; \cref{tab:native-err}).
Every failure falls into one of two categories:
\begin{itemize}[nosep,leftmargin=*]
\item \textbf{Silent wrong bits (\res{all}{native}{all}{silent} of \res{all}{native}{all}{err}, \res{all}{native}{all}{silentpct}\%).}
Every retokenized token lies inside the candidate set at its step, so the decoder terminates cleanly and returns bits that do not match the payload.
Without an external integrity tag, the receiver cannot tell that desynchronization occurred and therefore cannot request retransmission.
At the full vocabulary, almost all failures are silent (\res{llama}{native}{Full}{silent}, \res{qwen}{native}{Full}{silent} and \res{llmjp}{native}{Full}{silent} on English, Chinese and Japanese).
\item \textbf{Aborts outside the candidate set (\res{all}{native}{all}{oop} of \res{all}{native}{all}{err}).}
Canonical retokenization merges or splits emitted bytes into a token outside the candidate set at that step, halting the decoder.
These failures occur at every $k$ up to 4096; at the full vocabulary one remains, on Japanese.
\end{itemize}
In one Chinese and one Japanese text, the tokenizer merged the end of the prompt with the start of the text; Native's receiver then tokenizes the text alone (\cref{app:details}), which did not restore the sampled tokens there, and both texts are among the failures above.
Coin projection and SyncPool each decoded all \res{all}{recvall}{coin}{decoded} of their texts without error, and LAS, Stepwise, MWIS, Basic and hub-drop every text their receivers checked.

\paragraph{Native fails on the affected texts.}
All failures come from one event: canonical retokenization at the receiver does not return the sampled tokens.
We retokenize every text that Native, SyncPool and coin projection generate in \cref{tab:main,tab:e1-rest}, and call a text \emph{affected} when its canonical tokens after the prompt differ from the sampled ones, or when it merges with the prompt and tokenizing it alone does not restore them.
Every Native text that is not affected decodes correctly, and every affected one decodes wrongly except \res{all}{native}{all}{affectedok}.
Four of these are one memorized boilerplate passage on English, sampled at four values of $k$, whose 128 tokens carry so little entropy that no payload bit is fixed; the fifth is a Japanese text whose canonical tokenization splits a run of zeros differently, which Native's receiver still decodes correctly.
SyncPool's and coin projection's texts are affected too, but their receivers do not retokenize, and they decode correctly.

\paragraph{How rare ambiguity is.}
To compare with the measure of \citet{yan2025addressing}, we count the tokens of prompt and text whose byte span is not a span of the other tokenization (their SIT and CIT counts, on byte rather than character spans, so that tokens splitting a multi-byte character are counted too).
Over \res{llama}{amb}{all}{tokens}, \res{qwen}{amb}{all}{tokens} and \res{llmjp}{amb}{all}{tokens} generated tokens, this gives \res{llama}{amb}{all}{per1k}, \res{qwen}{amb}{all}{per1k} and \res{llmjp}{amb}{all}{per1k} tokens per 1,000 on English, Chinese and Japanese, and at most \res{all}{amb}{any}{max} in any setting of method and $k$ (\cref{sec:cost}).

\begin{table*}[t]
\centering
\footnotesize
\setlength{\tabcolsep}{4pt}
\begin{tabular}{@{}l rr rr rr@{}}
\toprule
 & \multicolumn{2}{c}{EN (Llama-3.1-8B)} & \multicolumn{2}{c}{ZH (Qwen3-8B-Base)} & \multicolumn{2}{c}{JA (llm-jp-3-7.2b)} \\
\cmidrule(lr){2-3}\cmidrule(lr){4-5}\cmidrule(lr){6-7}
$k$ & Silent & Pool & Silent & Pool & Silent & Pool \\
\midrule
16 & 2 & 0 & 1 & 2 & 3 & 4 \\
64 & 2 & 1 & 3 & 1 & 3 & 3 \\
256 & 4 & 0 & 7 & 5 & 9 & 2 \\
1024 & 4 & 1 & 15 & 1 & 9 & 3 \\
4096 & 5 & 0 & 14 & 2 & 16 & 3 \\
Full & 4 & 0 & 16 & 0 & 16 & 1 \\
\midrule
Total ($n=600$) & 21 & 2 & 56 & 11 & 56 & 16 \\
\bottomrule
\end{tabular}

\caption{Most of Native's extraction failures return wrong payload bits without raising an error.
Counts are over \res{all}{e1}{any}{nmin} paired texts per $(k,\text{language})$ cell after the text-alone fallback for prompt-boundary merges (\cref{app:details}).
\emph{Silent}: completed decoding with wrong bits; \emph{Pool}: retokenized token outside the candidate set.}
\label{tab:native-err}
\end{table*}

\section{Where the Prefix Forest Comes From}
\label{app:origin}

The hubs and lexical prefixes of \cref{sec:cost} are built into each vocabulary by the way its tokenizer is trained; \cref{tab:tokorigin} counts them from the tokenizer files alone.

\begin{table*}[t]
\centering
\small
\setlength{\tabcolsep}{6pt}
\begin{tabular}{@{}lrrr@{}}
\toprule
 & EN & ZH & JA \\
\midrule
Vocabulary construction & byte-level BPE & byte-level BPE & Unigram, byte fallback \\
Distinct byte strings & 128,000 & 151,643 & 99,441 \\
Single-byte strings & 256 & 256 & 256 \\
Begin with a space & 45.2\% & 35.0\% & 43.5\% \\
End inside a character & 1,068 & 1,255 & 51 \\
Merges whose left part is a token & 100\% & 100\% & -- \\
Han/kana words with a one-character prefix & 2,098 / 2,105 & 18,246 / 18,246 & 24,700 / 24,701 \\
Nodes (strings with an extension) & 41,302 & 45,399 & 23,850 \\
Internal levels & 85 & 85 & 15 \\
Extensions of the bare space & 57,874 & 53,020 & 43,266 \\
Largest lead-byte subtree & \texttt{0xE0}: 1,940 & \texttt{0xE5}: 7,012 & \texttt{0xE3}: 12,433 \\
\bottomrule
\end{tabular}

\caption{Every vocabulary nests short strings inside long ones, through merges in byte-level BPE and through frequent short forms in the Unigram vocabulary.
Counts are over distinct non-empty byte strings of regular tokens; a node is a string that is a proper prefix of another string; Han/kana words are strings of at least two Han or kana characters.}
\label{tab:tokorigin}
\end{table*}

\paragraph{Byte-level BPE (English and Chinese).}
The Llama-3 and Qwen3 tokenizers start from the 256 bytes and repeatedly merge two adjacent tokens into a new one \citep{sennrich2015neural,radford2019language}.
Two properties of the forest follow.
Every byte is a token, so at the full vocabulary every string lies in the tree of its first byte.
Every merge joins two existing tokens, so the left part of a merged token is itself a token; this holds for all \res{llama}{tok}{any}{merges} and \res{qwen}{tok}{any}{merges} merges, and it nests strings up to \res{llama}{tok}{any}{levels} levels deep.
Before merging, a regular expression splits the text into pre-tokens and attaches at most one preceding non-letter character, usually a space, to each run of letters.
Words are therefore learned together with their leading space: \res{llama}{tok}{any}{spaceshare}\% and \res{qwen}{tok}{any}{spaceshare}\% of the strings begin with one, and every such string extends the bare space token, which becomes the hub of English.
A Han character takes three bytes in UTF-8, and merges build it from the bottom: first partial characters (\res{qwen}{tok}{any}{partial} strings of Qwen3 end inside a character), then characters, then words.
Because the regular expression keeps a run of Han characters together as one pre-token, words are merged from their characters, and all \res{qwen}{tok}{any}{multi} Han/kana words of Qwen3 have a one-character prefix in the vocabulary.
The lead byte \texttt{0xE5} begins \res{qwen}{tok}{any}{leadext} strings.

\paragraph{Unigram with byte fallback (Japanese).}
llm-jp-3 uses a SentencePiece Unigram model \citep{kudo2018subword,kudo2018sentencepiece}, which starts from a large set of frequent substrings and prunes those whose removal costs the least likelihood.
Nothing in this construction requires the prefixes of a token to be tokens, and the forest is shallow, with \res{llmjp}{tok}{any}{levels} internal levels against \res{llama}{tok}{any}{levels}.
Prefixes still appear wherever a short form and its extensions are both frequent: \res{llmjp}{tok}{any}{multipref} of the \res{llmjp}{tok}{any}{multi} Han/kana words have a one-character prefix, and particles such as \emph{no} and \emph{ni} also begin compound particles.
SentencePiece splits the text at spaces and marks a word-initial space with ``\texttt{\spbar}'', so \res{llmjp}{tok}{any}{spaceshare}\% of the strings begin with a space.
Byte fallback adds the 256 bytes as tokens.
All kana and the CJK punctuation marks (\texttt{U+3000}--\texttt{U+303F}) begin with the lead byte \texttt{0xE3}, which therefore prefixes \res{llmjp}{tok}{any}{leadext} strings, \res{llmjp}{tok}{any}{leadshare}\% of the vocabulary.

\paragraph{Why ambiguous text is rare but the stop entropy is not.}
A tokenizer maps every training text to a single canonical tokenization.
The model therefore sees \texttt{\textvisiblespace weekend} as one token and almost never sees \texttt{\textvisiblespace week} followed by \texttt{end}, so it gives little probability to a continuation that spells a longer candidate; this is why non-canonical text is rare (\cref{app:native-err}).
The stop decision is a different quantity.
At a step where \texttt{\textvisiblespace week} and \texttt{\textvisiblespace weekend} are both likely, whether the token ends at \texttt{\textvisiblespace week} is uncertain before any continuation is drawn, and this uncertainty is $\hb(q_v)$.
Hubs are at the other extreme. The model almost never emits a bare space before a word or a lone lead byte, so $q_v\approx 0$, and a hub adds little to $\Hstop$ while it merges a large subtree into one root (\cref{sec:cost}); on English, the bare space alone produces \res{llama}{e0}{any}{spacebranchmin}--\res{llama}{e0}{any}{spacebranchmax}\% of $\Hbranch$ at every $k$.

\paragraph{Prompts in other languages.}
\Cref{tab:xlang} repeats the decomposition of \cref{fig:decomp} with each model on the prompts of all three languages.
Over the \res{all}{xlang}{any}{cells} pairs and every $k$, $\Hstop$ took \res{all}{xlang}{any}{stopmin}--\res{all}{xlang}{any}{stopmax}\% of $\Hstr$, so a model gives up a similar share on text it was not chosen for.
The split between root and branch follows the vocabulary instead. On Chinese at $k=256$, \res{llama}{xlang}{zh256}{branch}\% of $\Hstr$ lies in branches under Llama's byte-level BPE and \res{qwen}{xlang}{zh256}{branch}\% under Qwen's, whose Chinese characters mostly form separate roots.

\begin{table}[t]
\centering
\resizebox{\columnwidth}{!}{%
\footnotesize
\setlength{\tabcolsep}{3pt}
\begin{tabular}{@{}llrrrrrr@{}}
\toprule
 & & \multicolumn{3}{c}{$k=256$} & \multicolumn{3}{c}{Full} \\
\cmidrule(lr){3-5}\cmidrule(l){6-8}
Model & Prompts & Root & Branch & Stop & Root & Branch & Stop \\
\midrule
Llama-3.1-8B & EN$^\ast$ & 29.0 & 62.7 & 8.3 & 14.2 & 77.0 & 8.7 \\
 & ZH & 66.2 & 26.7 & 7.0 & 36.6 & 55.7 & 7.7 \\
 & JA & 59.7 & 32.1 & 8.2 & 31.6 & 59.2 & 9.2 \\
\addlinespace
Qwen3-8B-Base & EN & 32.1 & 59.8 & 8.1 & 15.4 & 76.0 & 8.6 \\
 & ZH$^\ast$ & 85.1 & 6.7 & 8.3 & 43.0 & 49.1 & 7.9 \\
 & JA & 79.0 & 12.1 & 8.8 & 34.7 & 55.8 & 9.5 \\
\addlinespace
llm-jp-3-7.2b & EN & 42.8 & 50.4 & 6.8 & 15.8 & 76.5 & 7.8 \\
 & ZH & 83.3 & 12.5 & 4.2 & 39.8 & 54.5 & 5.7 \\
 & JA$^\ast$ & 74.0 & 17.9 & 8.1 & 36.4 & 55.8 & 7.8 \\
\bottomrule
\end{tabular}
}
\caption{The stop share stays small on every model and language, while the split between root and branch depends on the vocabulary.
Shares of $\Hstr$ in \%, as in \cref{fig:decomp}; $^\ast$the language each model is evaluated on in \cref{sec:experiments}.}
\label{tab:xlang}
\end{table}

\section{Sender-Side Rejection}
\label{app:rejection}

Because most of Native's failures are silent (\cref{app:native-err}), the receiver cannot tell that it decoded the wrong bits, and asking for a resend would need a back channel anyway.
The sender can instead run the receiver on its own text and, on a failure, regenerate the text with fresh randomness (the receiver must then learn which attempt to decode).
This removes the failures but not the divergence.
If a text is regenerated with probability $r$, the accepted token sequences follow ordinary sampling conditioned on an event of probability $1-r$, and this law lies $-\log(1-r)$ nats from ordinary sampling.
By data processing, the law of the text lies at most this far, and in general it differs, because a text loses the probability of its non-canonical tokenizations and texts that merge with the prompt are removed altogether.
We estimate $r$ by the share of Native's texts that fail.
Over all vocabulary sizes, it is \res{llama}{rej}{128}{pct}\%, \res{qwen}{rej}{128}{pct}\% and \res{llmjp}{rej}{128}{pct}\% on English, Chinese and Japanese, so the text lies at most \res{llama}{rej}{128}{kl}, \res{qwen}{rej}{128}{kl} and \res{llmjp}{rej}{128}{kl} nats from ordinary sampling.
The rate grows with the length of the text, since every token is another chance for a merge. With $T=512$ (\cref{app:robust}), Native fails on \res{llama}{long}{Full}{nativeerr}, \res{qwen}{long}{Full}{nativeerr} and \res{llmjp}{long}{Full}{nativeerr} of \res{llama}{long}{Full}{n} texts per language at the full vocabulary, and regeneration would discard more than half of the Chinese and Japanese texts.
Stepwise applies the same idea at every step and pays the per-step divergences of \cref{tab:main}; summed over the at least 128 steps of a text, they exceed the per-text shift of regeneration for every model and $k$.
Coin projection needs no regeneration and keeps the law of ordinary sampling; what it pays instead is $\Hstop$.

\section{Multi-History Schemes}
\label{app:las}

Single-history schemes evaluate the language model only on prefixes of the sender's emitted token sequence $X_{<t}$.
Multi-history schemes such as LAS \citep{feng2025high} drop this restriction by letting sender and receiver keep the same weighted candidate token histories consistent with the visible text and query the model on more than one of them per step.
Because both sides can follow a string and its extensions at once, the message may also make the stop decisions.
In our runs (\cref{tab:main}), LAS turned \res{llama}{las}{Full}{utilpct}\%, \res{qwen}{las}{Full}{utilpct}\% and \res{llmjp}{las}{Full}{utilpct}\% of the full-model entropy into payload bits at the full vocabulary on English, Chinese and Japanese and decoded every checked text, with \res{llama}{time@las}{Full}{fwdpertok}, \res{qwen}{time@las}{Full}{fwdpertok} and \res{llmjp}{time@las}{Full}{fwdpertok} forward passes per token at the sender, up from \res{llama}{time@las}{16}{fwdpertok}, \res{qwen}{time@las}{16}{fwdpertok} and \res{llmjp}{time@las}{16}{fwdpertok} at $k=16$ (\cref{tab:timing}).
Its extra calls grow with the candidate set, while coin projection makes one call per token at every $k$.
LAS's utilization is close to the most any receiver of text can decode, since the text is a function of the tokens and so reveals at most their entropy, $\Hstr+\Hhom$ per step on average; here $\Hhom<10^{-4}$ bits (\cref{fig:decomp}).
By \cref{cor:budget}, a scheme that is exact and zero-error for every full-support model must call the model off the emitted history to recover any part of $\Hstop$.
Extra histories at the receiver alone would not suffice, since by the proof of \cref{thm:antichain} no receiver separates the candidates of a sender that calls the model only along its own history.

Our adapter runs LAS with its threshold $\tau=0$, at which it never falls back to synchronizing a whole group as SyncPool does, and runs its forward passes one at a time along a single cache, as the release does; batching them would shorten its time but not the number of calls, which is why we report both.
The zero in LAS's $\KL_{\mathrm{sec}}$ column is the exactness that \citet{feng2025high} state for it.
Their guarantee holds for every finite token source in which an end marker ends each history; our length rule is such a source, with the marker forced once a history has at least $T$ tokens and valid UTF-8 bytes, or $T+8$ tokens, so the guarantee carries over.
Our goodness-of-fit test (\cref{tab:gof}) covers Native, SyncPool and coin projection and does not cover LAS.
The two designs therefore answer different questions: LAS shows that nearly all of the string entropy can be recovered from text, whereas coin projection shows how much can be recovered at one model call per token and gives up $\Hstop$ (\cref{fig:decomp}) and nothing more.

\section{Repairing Errors: ReTokSync}
\label{app:retoksync}

\begin{table*}[t]
\centering
\footnotesize
\setlength{\tabcolsep}{3.5pt}
\begin{tabular}{@{}ll ccc c cc ccc c@{}}
\toprule
 & & \multicolumn{3}{c}{Util$\uparrow$} & ReTokSync / & \multicolumn{2}{c}{Texts with errors (\%)} & BER & Repair & Resets & \\
\cmidrule(lr){3-5}\cmidrule(lr){7-8}
Lang & $k$ & Native$^\ast$ & ReTokSync & Coin & Coin & Native$^\ast$ & ReTokSync & (\%) & (\%) & /1k tok. & $D$ \\
\midrule
 \multirow{2}{*}{EN} & 256 & .807 & .802 & .731 & 1.10\,{\scriptsize[1.05, 1.14]} & 2 & 4 & 0.11 & 0.66 & 17.2 & .068 \\
  & Full & .894 & .884 & .804 & 1.10\,{\scriptsize[1.07, 1.14]} & 8 & 6 & 0.10 & 0.53 & 4.4 & .046 \\
\midrule
 \multirow{2}{*}{ZH} & 256 & .786 & .788 & .722 & 1.09\,{\scriptsize[1.06, 1.12]} & 10 & 16 & 0.22 & 1.37 & 5.5 & .076 \\
  & Full & .918 & .921 & .841 & 1.09\,{\scriptsize[1.06, 1.13]} & 6 & 14 & 0.14 & 0.85 & 2.7 & .027 \\
\midrule
 \multirow{2}{*}{JA} & 256 & .761 & .757 & .696 & 1.09\,{\scriptsize[1.05, 1.13]} & 10 & 6 & 0.04 & 0.52 & 1.4 & .0053 \\
  & Full & .920 & .921 & .827 & 1.11\,{\scriptsize[1.07, 1.17]} & 18 & 16 & 0.14 & 0.84 & 1.9 & .0025 \\
\bottomrule
\end{tabular}
{R11 pending}}
\caption{ReTokSync codes nearly as much as Native under Discop, and its errors are a few wrong bits that a second text repairs rather than lost texts.
Texts of \cref{tab:coders}, \res{all}{rts}{any}{n} per cell; Util as in \cref{tab:main}; ReTokSync / Coin: ratio of the two utilizations with its 95\% paired bootstrap interval.
Texts with errors: texts whose decoded bits differ from the payload, for ReTokSync before repair; BER over ReTokSync's payload bits; Repair: correction message as a share of the payload; Resets per 1,000 tokens; $D$: per-step KL in nats between the next-token laws given the receiver's tokenization and given the emitted tokens, without truncation and averaged over all steps ($\KL_{\mathrm{sec}}$ at the full vocabulary).
$^\ast$Native's receiver tokenizes $\text{prompt}+\text{text}$.}
\label{tab:rts}
\end{table*}

\paragraph{Design.}
ReTokSync \citep{wang2026retoksync} conditions every step on the receiver's view, the canonical tokenization of the visible text.
After each token it retokenizes the text, and when the view changes it resets the token sequence, the generator state and the message pointer to what the receiver decodes from the new view.
When retokenization merges emitted tokens, the reset leaves a few wrong bits instead of a desynchronization.
A second channel repairs these bits: after every ten texts, SyncPool at $k=64$ carries one correction text that lists the position of each wrong token and the bits it should carry.
The coder thus receives the whole truncated law, as under Native, and Native's lost texts become local errors.

\paragraph{Implementation.}
No implementation had been released, so we implemented ReTokSync from its paper and supplement on our Discop port (\cref{app:coders}).
Our Discop draws the uniforms of step $t$ from the key and $t$, so resetting the generator state sets $t$ to the receiver's position.
The view is the canonical tokenization of the prompt and the text, as for Native's receiver, and is compared only when the text ends on a character boundary; a view token outside the candidate set carries no bits, for sender and receiver alike.
We count the bits of each correction message (an Elias-gamma count of the wrong tokens, then a position and the correct bits for each) instead of generating the correction text.
The receiver runs separately and rebuilds the view from the prompt; on every text it decoded exactly the bits that the sender predicted.

\paragraph{Why Discop.}
The range coder of \cref{tab:main} has no message pointer. It recovers the payload only by undoing every step's rotation after the last token, so ReTokSync's reset is not defined for it.
The supplement of \citet{wang2026retoksync} sketches ports to other coders, for arithmetic coding by also restoring the interval.
Rather than design a port to the range coder ourselves, we run ReTokSync on Discop, the coder of its main experiments, and pair it with Native and coin projection under Discop on the texts of \cref{tab:coders}, with the same prompts, keys and streams (\cref{tab:rts}).

\paragraph{Results.}
ReTokSync's utilization was \res{all}{rts}{any}{rtsnativemin}--\res{all}{rts}{any}{rtsnativemax}\% of Native's and \res{all}{rts}{any}{rtscoinmin}--\res{all}{rts}{any}{rtscoinmax} times coin projection's.
Native lost \res{all}{rts}{any}{nativefailmin}--\res{all}{rts}{any}{nativefailmax}\% of its texts; \res{all}{rts}{any}{texterrmin}--\res{all}{rts}{any}{texterrmax}\% of ReTokSync's texts carried wrong bits, but only a few each, at a bit error rate of at most \res{all}{rts}{any}{bermax}\%.
The correction messages took \res{all}{rts}{any}{corrratiomin}--\res{all}{rts}{any}{corrratiomax}\% of the payload.
Carried by SyncPool at $k=64$ at its rate in \cref{tab:e1-rest}, they add \res{all}{rts}{any}{extratokmin}--\res{all}{rts}{any}{extratokmax}\% tokens; counting these tokens, ReTokSync's utilization is \res{all}{rts}{any}{rtscoine2emin}--\res{all}{rts}{any}{rtscoine2emax} times coin projection's.
Its resets cost \res{all}{rts}{any}{callsmin}--\res{all}{rts}{any}{callsmax} model calls per token, plus a retokenization of the text at every step.

\paragraph{Two covers.}
ReTokSync's security analysis uses a cover that retokenizes the visible text $y$ before every step and samples the next token from $p(\cdot\mid\mathrm{Tok}(y))$; against this cover its visible text is exact \citep{wang2026retoksync}.
This is the natural cover when a model reads its own output back as text.
Ours, the cover of \cref{tab:main}, is ordinary sampling, which conditions on the tokens actually emitted.
The two agree while the emitted tokens are the canonical tokenization of the text and differ after a merge, where ReTokSync's sender conditions on a tokenization that it did not emit.
Column $D$ of \cref{tab:rts} measures this difference exactly, from the two laws at every step.
A next-token test from fixed histories, as in \cref{tab:gof}, sees it only at histories after a merge, because wherever the emitted tokens are the canonical tokenization of the text, ReTokSync and Native have the same next-token law.
At $k=256$, the top-256 sets given the two tokenizations differed on \res{all}{rts}{256}{klinfsharemin}--\res{all}{rts}{256}{klinfsharemax}\% of the steps; on these steps ReTokSync can emit a token that top-$k$ sampling would not, and $\KL_{\mathrm{sec}}$ is infinite, although the text alone does not reveal which tokens were emitted.
Which cover a warden faces depends on how cover texts are produced; under ReTokSync's own cover its divergence is zero.

\paragraph{The bound under ReTokSync's cover.}
Call a model \emph{text-reading} if every token sequence gets the next-token law of the canonical tokenization of its text, $\mathrm{LM}(x)=\mathrm{LM}(\mathrm{Tok}(\mathrm{Detok}(x)))$; for such models, ReTokSync's cover is ordinary sampling.
\Cref{thm:antichain} still holds when the class is defined over text-reading full-support models, with (ii) read as calling the model only on token sequences whose text is a prefix of the emitted text, which for these models means the same.
In its proof (\cref{app:proofs}), $\mathrm{LM}^\ast$ gives the branches $(h,x)$ and $(h,x')$ the same text and, at every common text, the same next-token law.
Indexing $\delta_j$ by the byte position after $\text{prompt}\cdot h\cdot\bytes(x)$ makes $\mathrm{LM}^\ast$ a function of the text, hence text-reading, and it agrees with $\mathrm{LM}_0$ on every text that does not extend $\text{prompt}\cdot h\cdot\bytes(x)$, including every prefix of $(\text{prompt},h)$.
The rest of the proof is unchanged, so \cref{thm:bound} and \cref{cor:budget} hold as well, and any scheme that is exact against ReTokSync's cover for every full-support model and has zero error still pays $\Hstop$.
ReTokSync goes beyond this budget because a merge makes its receiver decode wrong bits, which only the correction text, outside the single text that the bound concerns, can fix.

\paragraph{Reused uniforms.}
When a merge shortens the view, the reset moves the generator back to a step whose uniforms were already spent on an emitted token, and the next token reuses them.
Conditioned on the merge, these uniforms are no longer uniform, so that step departs slightly from both covers; this happened \res{all}{rts}{any}{reusemin}--\res{all}{rts}{any}{reusemax} times per 1,000 tokens, most often on English at $k=256$, where one text continues a table and each space it emitted merged with the run of spaces before it.
Drawing each uniform from the key and the byte offset at which its token starts would avoid the reuse, since every new token starts at the end of the text; we kept the reset as described.

ReTokSync hands the coder nearly the whole law and pays with errors that a second text repairs, under a cover that retokenizes; coin projection decodes every text on its own, exactly against ordinary sampling, and pays $\Hstop$.

\end{document}